\documentclass[10pt,twocolumn,letterpaper]{article}
\usepackage{booktabs}
\usepackage{siunitx}
\usepackage{multirow}
\usepackage{dblfloatfix}
\usepackage{pgfplots}
\usepackage{subcaption}
\pgfplotsset{compat=1.18}
\usepackage{algorithm}
\usepackage{algorithmic}
\usepackage[accsupp]{axessibility}
\usepackage{booktabs}
\usepackage{siunitx}
\usepackage{colortbl,xcolor}
\usepackage[dvipsnames]{xcolor}
\usepackage{float} 
\usepackage{booktabs,threeparttable}
\newcommand{\NA}{\textemdash} 
\definecolor{tblhead}{gray}{0.93} 
\definecolor{tblhilite}{gray}{0.90}

\usepackage{cvpr}      % To produce the REVIEW version
\newcommand{\append}[1]{Appendix \ref{#1}\xspace}
\definecolor{cvprblue}{rgb}{0.21,0.49,0.74}
\usepackage[pagebackref,breaklinks,colorlinks,allcolors=cvprblue]{hyperref}
\def\confName{CVPR}
\def\confYear{2026}

\usepackage{xspace}
\usepackage{todonotes}
\usepackage{xcolor}

\newcommand{\method}{UniGame\xspace}

\definecolor{szlpurple}{RGB}{106,13,173} % 可按需改成你喜欢的紫色

\title{UniGame: Turning a Unified Multimodal Model Into Its Own Adversary}

\author{
Zhaolong Su$^{1}$ \quad
Wang Lu$^{2}$ \quad
Hao Chen$^{3}$ \quad
Sharon Li$^{4}$ \quad
Jindong Wang$^{1}$\thanks{Corresponding author.}\\
$^{1}$William \& Mary
$^{2}$Independent
$^{3}$Carnegie Mellon University
$^{4}$University of Wisconsin--Madison\\
{\tt\small \{zsu05,jdw\}@wm.edu, newlw230630@gmail.com, haoc3@andrew.cmu.edu, sharonli@cs.wisc.edu}
}

\begin{document}
\maketitle

\begin{abstract}
Unified Multimodal Models (UMMs) have shown impressive performance in both understanding and generation with a single architecture. However, UMMs still exhibit a fundamental inconsistency: understanding favors compact embeddings, whereas generation favors reconstruction-rich representations.
This structural trade-off produces misaligned decision boundaries, degraded cross-modal coherence, and heightened vulnerability under distributional and adversarial shifts.
In this paper, we present \textbf{\method}, a self-adversarial post-training framework that directly targets the inconsistencies.
By applying a lightweight perturber at the shared token interface, \method enables the generation branch to actively \textit{seek and challenge} fragile understanding, turning the model itself into its own adversary.
Experiments demonstrate that \method significantly improves the consistency ($+4.6\%$).
Moreover, it also achieves substantial improvements in understanding ($+3.6\%$), generation ($+0.02$) on GenEval, out-of-distribution and adversarial robustness ($+4.8\%$ and $+6.2\%$ on NaturalBench and AdVQA).
The framework is architecture-agnostic, introduces $<1\%$ additional parameters, and is complementary to existing post-training methods. These results position adversarial self-play as a general and effective principle for enhancing the coherence, stability, and unified competence of future multimodal models.
Code is available at \url{https://github.com/AIFrontierLab/TorchUMM}.

\end{abstract}

\section{Introduction}
\label{sec:intro}

\begin{figure*}[t]
  \centering
  \captionsetup[sub]{font=scriptsize}
  \begin{subfigure}[t]{0.24\textwidth}
    \centering
    \includegraphics[width=\linewidth]{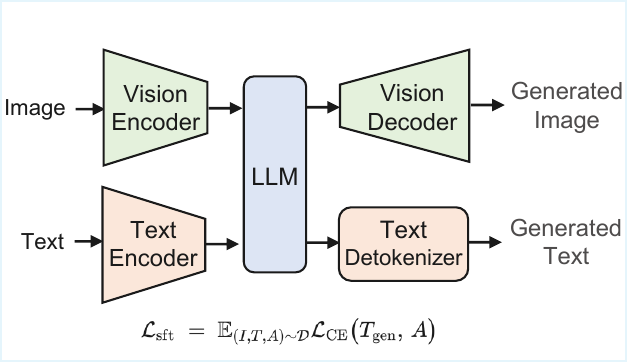}
    \caption{Supervised fine-tuning\\}
    \label{fig-compare-sft}
  \end{subfigure}\hfill
  \begin{subfigure}[t]{0.24\textwidth}
    \centering
    \captionsetup[sub]{font=scriptsize}
    \includegraphics[width=\linewidth]{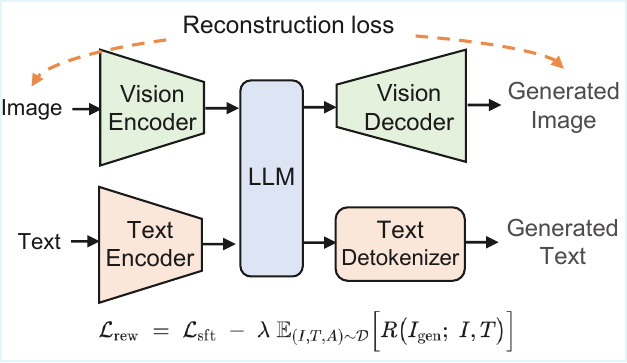}
    \caption{Reconstruction-base approaches\\
   }
    \label{fig-compare-recon}
  \end{subfigure}\hfill
  \begin{subfigure}[t]{0.24\textwidth}
    \centering
    \captionsetup[sub]{font=scriptsize}
    \includegraphics[width=\linewidth]{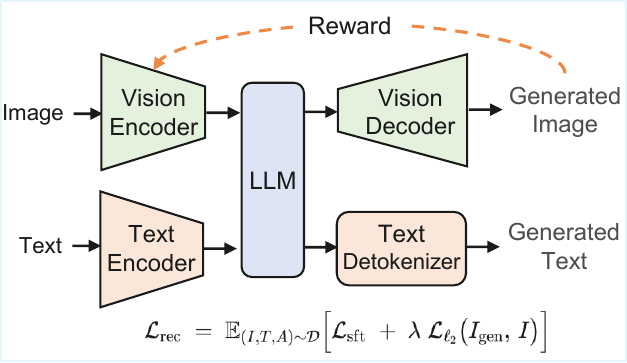}
    \caption{Reward-based approaches\\
   }
    \label{fig-compare-reward}
  \end{subfigure}\hfill
  \begin{subfigure}[t]{0.245\textwidth}
    \centering
    \captionsetup[sub]{font=scriptsize}
    \includegraphics[width=\linewidth]{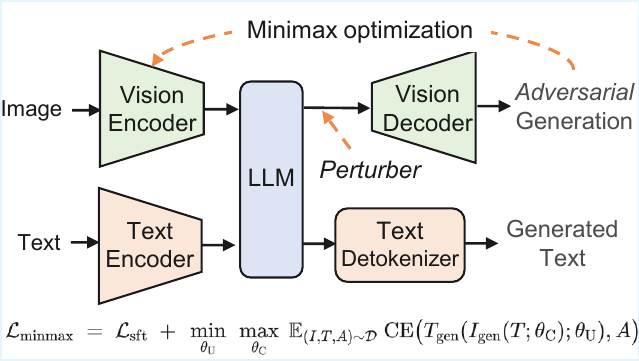}
    \caption{\emph{\textbf{Ours}}: Minmax optimization\\
    }
    \label{fig-compare-ours}
  \end{subfigure}
  \vspace{-.1in}
  \caption{Illustration of four different post-training paradigms.}
  \label{fig-compare}
  \vspace{-.2in}
\end{figure*}

Unified Multimodal Models (UMMs) have recently demonstrated impressive capability in both visual understanding and image generation with a unified architecture~\cite{deng2025emerging,wang2024emu3,qu2025tokenflow,xie2024show,xiao2025omnigen,wu2025harmonizing}.
By jointly leveraging a language model backbone and a visual tokenizer–decoder stack~\citep{li2025dual,chen2024janus}, these models promise a unified interface for cross-modal reasoning, grounded perception, and controllable generation.
Specifically, the large-scale pre-training establishes general multimodal capabilities, and the post-training stage (supervised fine-tuning, SFT, \Cref{fig-compare-sft}) can further improve their performance on downstream tasks with enhanced reliability.
\begin{figure}[t]
  \centering
  \begin{subfigure}[b]{0.51\linewidth}
    \centering
    \includegraphics[width=\linewidth]{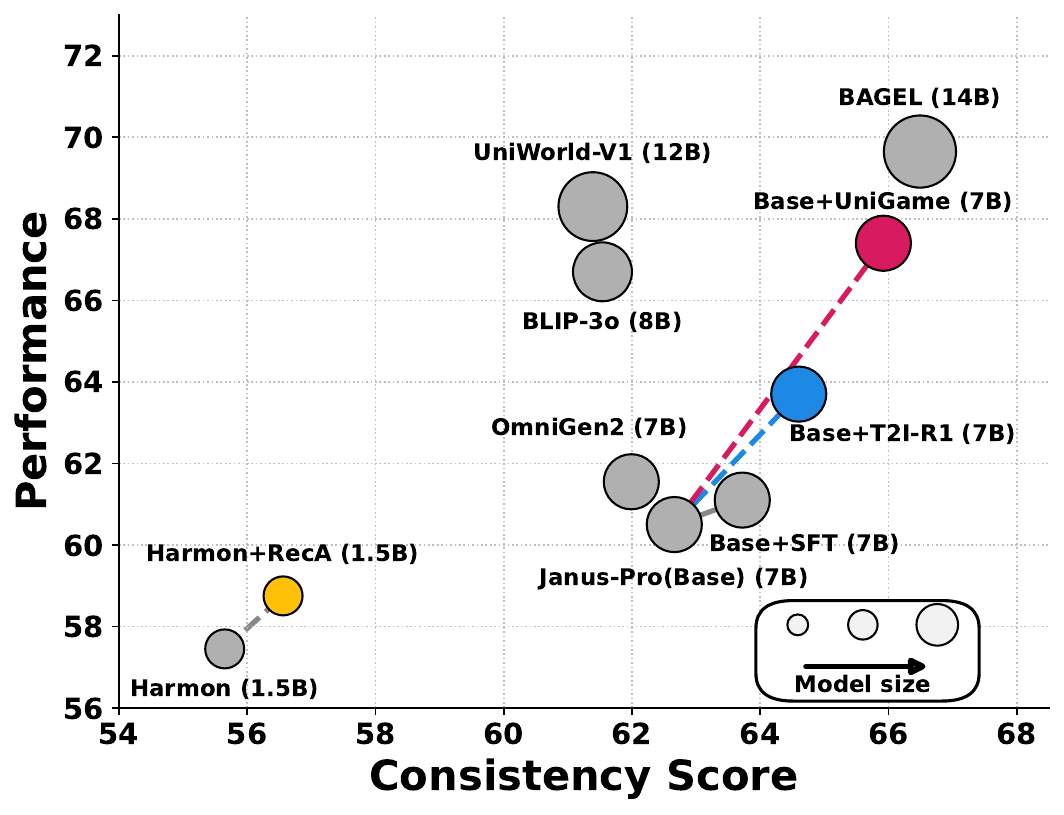}
    \caption{Performance vs. consistency.}
    \label{fig-p-v-c}
  \end{subfigure}
  \hfill
  \begin{subfigure}[b]{0.48\linewidth}
    \centering
    \includegraphics[width=\linewidth]{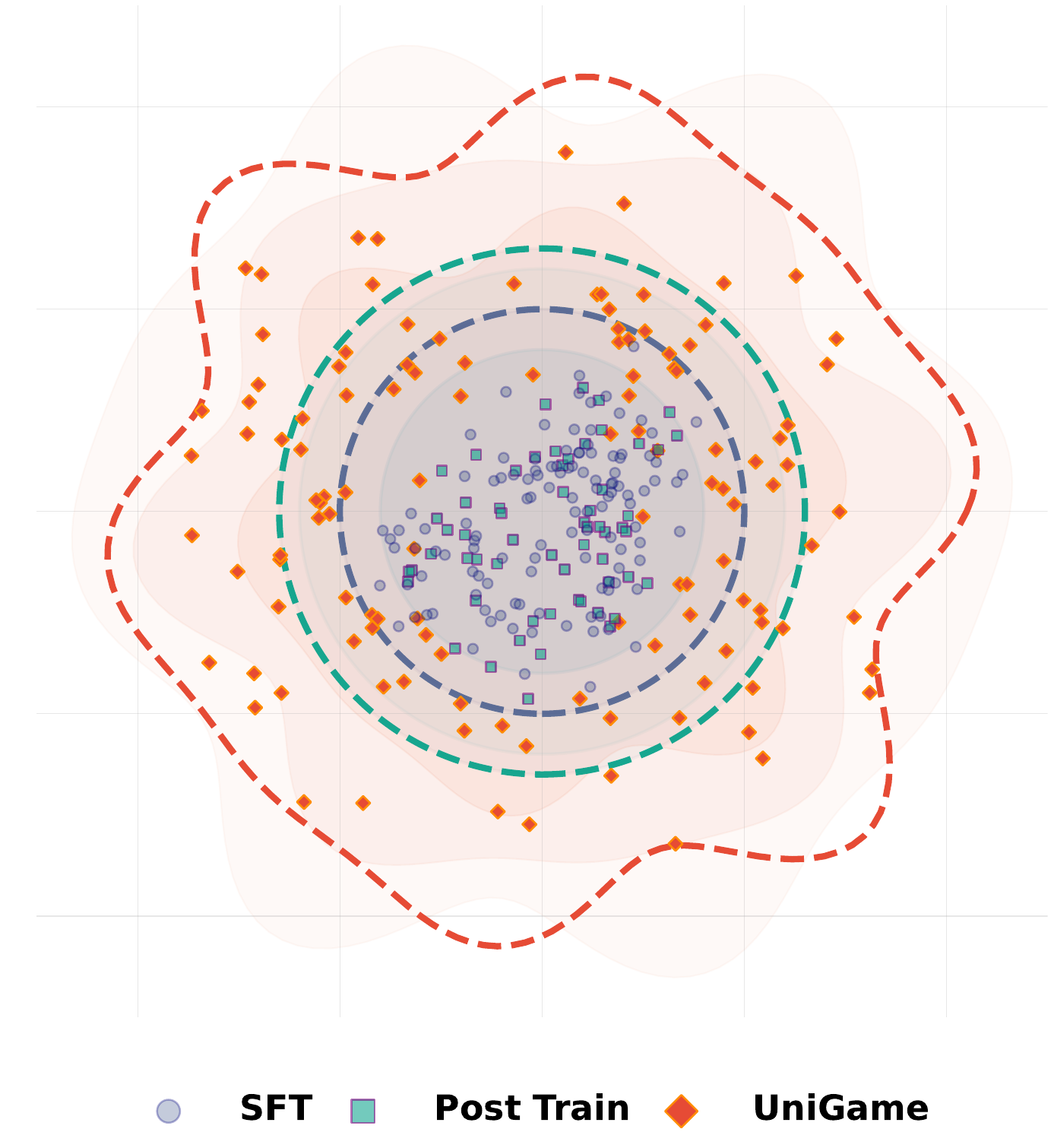}
    \caption{Manifold coverage.}
    \label{fig:manifold_coverage}
  \end{subfigure}
  \caption{\textbf{Qualitative and quantitative analyses of \method.\protect\footnotemark}
  (a) The performance vs. consistency score of several models, indicating significant improvement of both metrics of our models.
  (b) The manifold produced by SFT, reconstruction-based Post Train, and \method. \method expands the training distribution toward hard yet realistic neighborhoods.}
  \label{fig:manifold}
  \vspace{-.2in}
\end{figure}
\footnotetext{(a) Where consistency Score computed as the average of WISE and UnifiedBench, performance is averaged over understanding bench MMMU and generation bench like GenEval. (b) We randomly sample 100 images, extract their unified embeddings, project to 2D with UMAP~\citep{mcinnes2018umap}; colored regions visualize each method’s \emph{on-manifold} coverage. }

%  $\boxed{\;
% \mathcal{L}_{\mathrm{sft}}
% \;=\;
% \mathbb{E}_{(I,T,A)\sim\mathcal{D}} 
% \mathcal{L}_{\mathrm{CE}}\big(T_{\mathrm{gen}},\,A\big)
% \;}$
%  $\boxed{\;
% \mathcal{L}_{\mathrm{rec}}
% \;=\;
% \mathbb{E}_{(I,T,A)\sim\mathcal{D}}\Big[
% \mathcal{L}_{\mathrm{sft}}
% \;+\;
% \lambda\;\mathcal{L}_{\ell_2}\big(I_{\mathrm{gen}},\,I\big)
% \Big]
% \;}$

%  $
%     \boxed{\;
% \mathcal{L}_{\mathrm{rew}}
% \;=\;
% \mathcal{L}_{\mathrm{sft}}
% \;-\;
% \lambda\;
% \mathbb{E}_{(I,T,A)\sim\mathcal{D}}\Big[
% R\big(I_{\mathrm{gen}};\;I,T\big)
% \Big]
% \;}
%     $

%     $\boxed{\;
% \mathcal{L}_{\mathrm{minmax}}
% \;=\;
% \mathcal{L}_{\mathrm{sft}}
% \;+\;
% \min_{\theta_{\mathrm{U}}}\;\max_{\theta_{\mathrm{C}}}
% \;
% \mathbb{E}_{(I,T,A)\sim\mathcal{D}}\;
% \mathrm{CE}\big(T_{\mathrm{gen}}(I_{\mathrm{gen}}(T;\theta_{\mathrm{C}});\theta_{\mathrm{U}}),A\big)
% \;}$

Despite their great performance, UMMs exhibit structural \emph{inconsistency} between their understanding and generation pathways~\citep{yan2025can,wang2024reconstruction}.
This inconsistency stems from the inherently conflicting nature of the two objectives, leading to the mismatch in various aspects such as \textit{semantics} (i.e., the model can answer a question correctly yet fail to generate a corresponding image, or vice versa \cite{Gan2020,rajabi2025token}), \textit{capability} (e.g., generation is harder to improve than understanding, or vice versa~\citep{radford2021learning}), and \textit{feature compactness} (e.g., understanding requires more compact feature space while generation prefers oppositely).
Inconsistency widely exists in real-world applications, where models frequently encounter unexpected inputs far from the training manifold, compositional combinations unseen during training, counterfactual queries, or modality conflicts such as distribution shift~\citep{Gan2020, li2024naturalbench, oh2025understanding, oh2025vittle} and adversarial attack~\citep{li2021adversarial}.
If not sufficiently studied, it would greatly undermine multimodal information fusion, model robustness, and further performance improvement.

It remains challenging and unexplored to improve the consistency of UMMs, primarily due to the unclear learning objectives: only the consequences, but not the causes, are known.
Therefore, recent post-training approaches tend to fill this gap using surrogate objectives.
Reconstruction-based approaches (\Cref{fig-compare-recon}) regenerate the original images through the semantic embedding space derived from visual perception~\citep{wang2024reconstruction,yan2025can}. 
They optimize a unified reconstruction objective, which trains models within a closed auto-encoding loop. 
Reward-based methods (\Cref{fig-compare-reward}) typically optimize an ensembled reward function \citep{jiang2025t2i}, combining task-specific or rule-based metrics to refine the output relying on external expert models.  
However, both are optimized using handcrafted objectives (reconstruction or reward), which only polish model behavior on a fixed training distribution and place no explicit constraints on the two coupling branches.
As a result, they reproduce behaviors within a fixed manifold rather than expanding the shared generative space, leaving the inconsistency largely unresolved.

\begin{figure*}[t]  
  \centering
  \includegraphics[width=.9\textwidth]{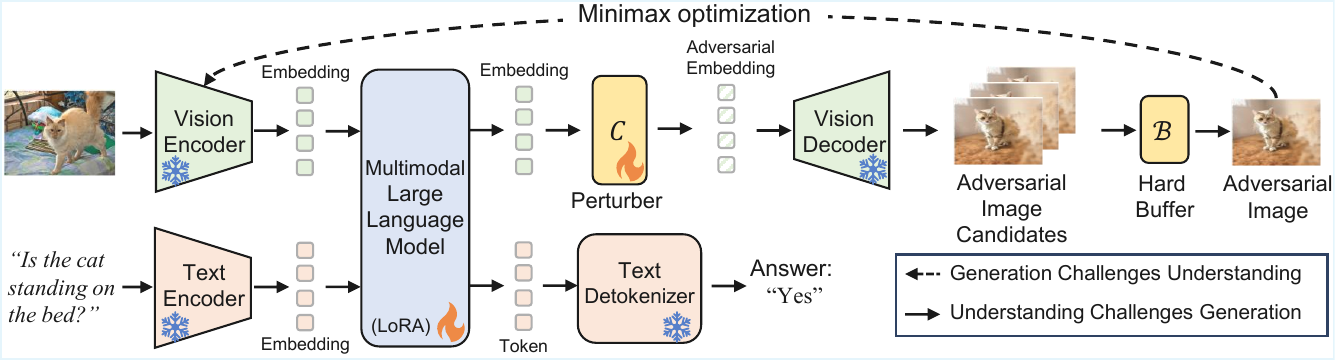}
  \vspace{-.1in}
  \caption{\textbf{Overview of \method.}
   This adversarial self-play improves understanding robustness and understanding-generation consistency. The perturber $C$ is a lightweight (3-layer MLP) module and the hard buffer $\mathcal{B}$ is a filtering mechanism.}
   \vspace{-.2in}
  \label{fig:method}
\end{figure*}

\emph{Can a UMM expose and correct its own inconsistencies from within?}
In this paper, motivated by the observation that adversarial signals reliably surface brittle reasoning in vision–language models~\citep{Gan2020,li2024naturalbench,oh2025understanding,rajabi2025token}, we propose \textbf{\method} (\Cref{fig:method}), the first self-adversarial post-training framework for UMMs.
\method treats the generation pathway as an active adversary that searches for visually plausible, decoder-constrained perturbations that maximally challenge the understanding branch.
It pushes the model beyond fixed data manifolds and produces structured adversarial samples along the uncertainty regions (\Cref{fig-p-v-c}).
Concretely, \method installs a lightweight perturber at the shared visual-token interface to create bounded, structured perturbations.
These perturbations are decoded into realistic adversarial images, filtered through a semantic consistency check, and stored in a hard-example buffer.
The understanding branch is then optimized to correctly reason over both clean inputs and these internally generated, semantically aligned counterexamples.
This forms a minimax self-play process where generation seeks to expose weaknesses while understanding learns from them, effectively expanding the shared generative manifold toward fragile yet meaningful regions (\Cref{fig:manifold_coverage}; theoretical insights in \append{sec-append-theory-mani}).
Compared to existing efforts, \method explicitly converts representational weaknesses into decoded, semantically coherent counterexamples that efficiently harden understanding (\Cref{fig-compare}). 
Empirically, \method uncovers richer, actionable failure modes (e.g., counting, fine attributes, and occlusion in \Cref{fig:casestudy}) and improves the consistency, performance, and robustness.

This paper makes the following contributions:
\begin{enumerate}
    \item \textbf{Novel framework.} We propose \method as the first framework to formalize UMMs post-training as a self-play game to improve the consistency between the understanding and generative pathways.
    \item \textbf{Self-play training algorithm.} We instantiate \method by devising a flexible co-training algorithm that combines the perturber, regularizer, and hardness-aware mining modules. The algorithm is agnostic to both UMM architectures and post-training approaches.
    \item \textbf{Empirical improvement.} \method yields improvements in consistency (4.6\%), understanding (+3.6\%), generation (+0.02 on GenEval), OOD (+4.8\%) and adversarial robustness (+6.2\%).
\end{enumerate}

\section{Related Work}
\label{sec-related}

% \textbf{Unified Multimodal Models.}
% \wjd{A brief introduction of UMMs}

% \textbf{Unified Training of UMMs.}
% \wjd{Survey existing self-improvement research.}
% \citet{yan2025can} proposed to formulate the understanding and generation objectives using the autoencoder-based reconstruction, which introduces extra modules such as CLIP~\citep{radford2021learning} semantic alignment, GPT-4o distillation, and Intern-VL filtering.
% In contrast, ours does not introduce new foundation models, which is conceptually simpler and more efficient.

\noindent \textbf{Unified Multimodal Models.}  
UMMs aim to combine multimodal understanding and generation within a single backbone, enabling compact deployment and richer cross-modal reasoning \cite{chen2025blip3, wang2024emu3, team2024chameleon,chen2025janus,qu2025tokenflow,deng2025emerging,qu2025tokenflow}.
Among these, BLIP3-o~\citep{chen2025blip3} explores hybrid autoregressive and diffusion training recipes to balance understanding and generation fidelity.
Emu3~\citep{wang2024emu3} treats images and text as an interleaved token stream and scales next-token prediction to unified multimodal outputs.
TokenFlow~\citep{qu2025tokenflow} focuses on the tokenizer layer and introduces dual-granularity codebooks to reconcile the conflicting demands of discriminative understanding vs reconstructive generation.
Despite architectural innovation, UMMs continue to face the inconsistency issue: the representation granularity and objective tension that underlie understanding and generation still cause ambiguous shared embeddings and latent failure modes that remain insufficiently addressed.

\noindent \textbf{Post-training of UMMs.}
Aiming to resolve the inconsistency issue, existing post-training methods can be categorized into these types: (i) reconstruction and semantic-alignment losses to encourage fidelity~\citep{wang2024reconstruction,yan2025can}; (ii) RL/reward-based optimization to directly optimize downstream metrics~\citep{Ramesh2024,jiang2025t2i}. 
Each family improves either fidelity or robustness~\citep{Li2023,chen2024janus,wang2024reconstruction,Gan2020,rajabi2025token,Ramesh2024,qu2025tokenflow}, For instance, RecA~\citep{wang2024reconstruction} leverages reconstruction alignment by conditioning generation on understanding embeddings and using reconstruction losses to bring representations closer, while T2I-R1~\citep{jiang2025t2i} enhances image generation through collaborative semantic- and token-level chain-of-thought combined with reinforcement learning. 
In the vision–language domain, AT has shown potential: VILLA introduces large-scale embedding-space perturbations across image and text modalities, improving robustness and generalization \cite{Gan2020,rajabi2025token}. 
Nevertheless, most methods either improve alignment or generation separately, and adversarial mechanisms are rarely incorporated into full UMMs. 
They commonly fail in one crucial respect: they do not exploit generation as an active adversarial process to strengthen the understanding branch. 
Specifically, adversarial or unguided embedding perturbations often produce off-manifold samples—unrealistic or semantically invalid artifacts, while reconstruction objectives do not intentionally surface decision-critical failure modes. 
Although RL-based schemes are effective, they are computationally costly and do not guarantee that these discovered examples remain decodable and semantically plausible~\citep{Ramesh2024,jiang2025t2i}. 
This leaves a key gap in truly improving the consistency of the understanding and generation pathways.

%%%%%%%%%%%%%%%%%%%%%%%%%%%%%%%%%%%%%%%%%%%%%%%%%%%%%%%%%%%%%%%%% section 3
\section{\method}
\label{sec:method}

% We refer the original generation branch as the \textit{clean} path, then, the perturber introduces a \emph{perturbed} path.
% In \method, a frozen visual encoder maps an input image to visual tokens at the unified interface.
% In the clean path, it is forwarded to the understanding head for supervised training with the query $q$;
% In the perturbed path, the perturbations generated by $C$ are decoded by $G$ into candidate images.
% The semantically consistent candidates are stored in the buffer $\mathcal{B}$.
% During training, the understanding module learns from clean samples and hard examples from $\mathcal{B}$, while $C$ is optimized to generate challenging yet plausible cases.
% During training, the vision encoder (SigLIP~\citep{xx}) is frozen, and only LoRA~\cite{xx} adapters on the LLM backbone plus the Perturber $C$ are trainable. 

% The overall training objective of \method is:
% \begin{equation}
% \label{eq:minmax}
% \begin{split}
% \min_{\theta_U}\;\max_{\theta_C}\; \mathbb{E}_{(\mathbf x,q,a)} \Big[ \;&
% \mathrm{CE}\big(p_U(a \mid C(\mathbf{z}; \theta_C),q;\theta_U),a\big) \\
% & - \lambda \|\boldsymbol\delta\|^2 \Big].
% \end{split}
% \end{equation}

% In practice, we (i) decode \(\tilde{\mathbf z}\)
% via \(G\) to obtain image candidates, (ii) , and (iii) re-encode or score decoded candidates for mining and replay.

\subsection{Preliminary}
\label{sec:prelim}

Let $\mathbf{x}$ denote an image, $q$ a vision-grounded query, and $a$ the ground-truth answer. 
After a frozen visual encoder and a projection layer, we write the unified visual tokenizer, which quantizes image representations into tokens aligned with the language model’s vocabulary embedding space as $\mathbf{z} = \mathrm{Enc}(\mathbf{x}) \in \mathbb{R}^{N \times H}$,
where $N$ is the token length and $H$ the hidden dimension.\footnote{In addition to encoder and projection layers, real encoder modules consists of other parts such as semantic encoder and tokenizer. We omit these details for simplicity.}
Both visual and textual embeddings are then fed into a language model to learn high-level embeddings.
The understanding branch $U$ aims to minimize the discriminative loss of the textual output, and the generation branch $G$ tends to reconstruct the input image: 
\begin{align}
    \mathcal{L}_{\mathrm{und}}(\theta_U) &:= -\mathbb{E}\big[\log p_U(a\mid\mathbf{z},q)\big], \\
    \mathcal{L}_{\mathrm{gen}}(\theta_G) &:= \mathbb{E}\big[\ell_{\mathrm{gen}}(G(\mathbf{z}),\mathbf{x})\big],
\end{align}
where $p_U(a\mid\mathbf{z},q)$ is the predictive distribution of the understanding task and $\ell_\text{gen}$ is the reconstruction loss (e.g. MSE).
These two objectives are commonly optimized jointly in a single multi-task objective:
\begin{equation}
\min_{\theta_U,\theta_G}\ \mathcal{L}_{\mathrm{joint}}
:=\mathcal{L}_{\mathrm{und}}(\theta_U)+\lambda\,\mathcal{L}_{\mathrm{gen}}(\theta_G),
\end{equation}
where \(\lambda\) is the trade-off hyperparameter.\footnote{The term “jointly” here denotes simultaneous (multi-task) optimization of both branches—possibly sharing backbone parameters—rather than strictly sequential stage-wise training.}
% While this joint objective is standard in existing UMMs, it treats the understanding and generation branches cooperatively and does not explicitly enforce their consistency.

\subsection{Motivation}
\label{sec-method-motiv}

UMMs inherently exhibit inconsistencies between the understanding and generation pathways due to their conflicting optimization requirements: the understanding branch favors task-oriented embeddings, but generation demands reconstruction-rich representations. 
Both branches operate on a shared generative manifold induced by encoding interface and decoder; any mismatch in how they carve up this manifold directly translates into structural inconsistencies. 
Improving consistency is challenging primarily owing to the lack of clear, direct learning objectives:
We can only observe the consequences (e.g., semantic mismatches, capability gaps) but struggle to identify their underlying causes.

Existing efforts~\citep{wang2024reconstruction,jiang2025t2i} optimize for individual goals on a fixed set of data distributions, and place no explicit constraints on the two coupling training objectives.
They encourage cooperative training, reproducing existing samples rather than expanding coverage of the shared manifold, where boundary behavior is most fragile, thus aggravating inconsistency.
We argue that improving consistency requires \emph{expanding} this shared manifold, especially around decision boundaries, instead of merely polishing the model within its comfort regions.

Considering the unified architecture of UMMs: can we improve the consistency \textit{within} the model itself?
We turn to adversarial training~\citep{Madry2017}, which creates adversarial perturbations to explore understanding failures~\cite{Gan2020,rajabi2025token}.
This indicates that adversarial signals, if properly constrained, can serve as an effective mechanism for regularizing the decision boundaries of UMMs.
Within the UMM architecture, we focus on converting generative priors into \emph{decodable adversarial cases} that (i) remain semantically valid and (ii) reliably expose genuine reasoning failures in the understanding branch.
This intuitively motivates the self-play training paradigm that makes best use of both of the understanding and generation branches as a minimax optimization framework.
The generative pathway no longer passively follows alignment objectives: it is explicitly trained to produce realistic, on-manifold adversarial cases that challenge the understanding module, 
while the understanding branch is optimized to solve these internally generated challenges. 

\subsection{Overview of \method}
\label{sec:framework}
As shown in \Cref{fig:method}, the proposed \textbf{\method} introduces two lightweight, plug-in modules to a general UMM:
\begin{itemize}
  \item \textbf{Perturber \(C\):} A compact network with
    parameters \(\theta_C\) ($|\theta_C| \ll \min(|\theta_U|,|\theta_G|)$) that maps the post-LM fused visual states \(\hat{\mathbf z}\) to a perturbed token: $\tilde{\mathbf z}=C(\hat{\mathbf{z}}; \theta_C)= \hat{\mathbf z}+\boldsymbol\delta$, where \(\|\boldsymbol\delta\|\le\varepsilon_{\max}\) is the budget to cap the perturbation magnitude for stabilization, We control the budget through an ablation study and use the best-performing setting in all main experiments. Details of the perturber architecture are in \append{perturber}.
  \item \textbf{Hard-sample buffer \(\mathcal B\):} A component that
    scores decoded candidates and stores hard, semantically plausible examples for replay via semantic-consistency check~\citep{radford2021learning}:
    \begin{equation}
    \label{eq:buffer}
    \mathcal{B} = \big\{\,G(\tilde{\mathbf z}) \;\big|\; H(\tilde{\mathbf z}) \ge \tau \,\big\},
\end{equation}
where $H = \mathrm{CE}\big(p_U(\hat{a}\mid \mathrm{Enc}(G(\tilde{\mathbf z})), q; \theta_U), a\big)$ is the cross-entropy loss and $\tau$ is the threshold.
\end{itemize}

We refer the original generation branch as the \textit{clean} path, then the perturber introduces a \emph{perturbed} path.
In \method, a frozen visual encoder maps an input image to visual tokens at the unified interface.
In the clean path, it is forwarded to the understanding head for supervised training with the query $q$;
In the perturbed path, the embeddings with perturbations generated by $C$ are decoded by $G$ into candidate images.
The semantically consistent candidates are stored in the buffer $\mathcal{B}$.
During training, the understanding module learns from hard examples from $\mathcal{B}$ and clean samples, and the perturber $C$ is optimized to generate challenging yet plausible cases.
The vision encoder (SigLIP~\citep{zhai2023sigmoid}) is frozen, and only LoRA~\cite{hu2022lora} adapters on the LLM backbone and the Perturber $C$ are trainable. 

The overall training objective is a minimax game:
\begin{equation}
\label{eq:minmax}
\min_{\theta_U}\max_{\theta_C}
\Big(
  \mathcal{L}_{\mathrm{U}}(\theta_U)
  + 
    \lambda \mathcal{L}_{\mathrm{C}}(\theta_C;\theta_U)
\Big),
\end{equation}
where $\lambda>0$ controls the strength of the self-play signal between these two branches.
We provide some \textbf{theoretical insights} to the convergence in \append{sec:theory}: Under bounded perturbation and decoder constraints, the perturber optimizes a lower bound of the worst-case understanding loss. This ensures the minimax dynamics remain stable and prevents off-manifold adversarial drift.
We will elaborate on $\mathcal{L}_{\mathrm{U}}$ and $\mathcal{L}_{\mathrm{C}}$ in next section.

\subsection{The Self-Play Training Process}
\label{sec-method-training}
The training objective in Eq.~\eqref{eq:minmax} consists primarily of two adversarial and iterative steps: to enable the understanding and generation branches to challenge each other.
The complete training procedure is presented in \append{sec-append-algo} and we introduce only these two challenging steps.

\noindent \textbf{Understanding Challenges Generation} (the solid arrows in \Cref{fig:method}).
This is the naive feedforward path that optimizes the understanding module to prevent the generation branch from confusing it, i.e., to ``challenge'' the generation branch.
The clean path forwards the original visual tokens \(\mathbf z\) directly to the understanding head \(U\), producing the model's nominal predictions and the standard supervised loss.

Here, the clean path preserves the original semantic information and simply computes the supervised loss; semantic plausibility and regularization are enforced on the generation side via the perturber’s norm and CLIP-based filtering from the hard-example mining buffer. Formally:
\begin{equation}
\label{eq:lossU}
\begin{split}
\mathcal{L}_U(\theta_U) =\; & 
\mathbb{E}_{\mathrm{clean}}\big[\mathrm{CE}(p_U(\hat{a}\mid\mathbf z,q;\theta_U),a)\big] \\
& + \beta\,\mathbb{E}_{\mathcal B}\big[\mathrm{CE}(p_U(\hat{a}\mid \mathbf{z},q;\theta_U),a)\big],
\end{split}
\end{equation}
where the first term keeps the model accurate on clean data, the second term forces $U$ to correctly answer on current adversarial examples and mined hard cases, and $\beta > 0$ is a trade-off hyperparameter.

\noindent \textbf{Generation Challenges Understanding} (the dashed arrow in \Cref{fig:method}).
In this process, the perturbed embedding $\tilde{\mathbf{z}}$ will be rendered by the decoder $G$ into image candidates: \(\tilde{\mathbf x}=G(\tilde{\mathbf z})\),
which are then subject to semantic-consistency checks (e.g., CLIP~\citep{radford2021learning} similarity)
and re-encoding/scoring by the understanding module. 
This path intentionally produces \emph{on-manifold} adversarial examples to challenge the understanding branch, 
and hard candidates are stored in the buffer \(\mathcal B\) for replay.
Formally, the perturber is updated to maximize the understanding loss:
\begin{equation}
\label{eq:lossC}
\begin{split}
&\mathcal{L}_C(\theta_C;\theta_U) =\;  \\
&\mathbb{E}_{\mathrm{clean}}\mathrm{CE}\big(p_U(\hat{a} \mid \mathrm{Enc}(G(C(\hat{\mathbf{z}}; \theta_C))),q;\theta_U),a\big)
- \lambda\|\boldsymbol\delta\|^2. \\
\end{split}
\end{equation}
\subsection{Discussion}
\textbf{\method vs. GANs.}
% \method is different from conventional GANs \citep{Goodfellow2014gan} that train a generator to fool a discriminator~\citep{Goodfellow2014gan,Radford2015}.
% First, GAN needs an extra discriminator for the adversarial game while ours can operate \textit{within} an UMM to leverage its own understanding and generation branches.
% Second, GANs primarily aim to improve image synthesis quality, while \method jointly targets generation and understanding consistency: the perturbation occurs in the shared visual-token space and decoded back to images, not by adding pixel noise that triger model's failure, this way, it exposes cross-modal reasoning failures, while gaining adversarial semantic information.
\method is different from conventional GANs \citep{Goodfellow2014gan} that train a generator to fool a discriminator~\citep{Goodfellow2014gan,Radford2015}. First, GAN needs an extra discriminator for the adversarial game, while ours can operate \textit{within} a UMM to leverage its own understanding and generation branches. Second, GANs primarily focus on generation tasks while ours targets both generation and understanding, involving more complex training and optimization process.

\noindent \textbf{\method vs. Adversarial Training (AT).}
\method is the first attempt of applying AT~\citep{Madry2017,Goodfellow2014adv} to UMMs, but has the following differences.
First, AT is mainly used for enhancing adversarial robustness, while ours explores AT for consistency improvement.
Second, \method differs fundamentally from prior AT by enforcing decoder-constrained, on-manifold image perturbations, enabling self-generated adversarial cases that remain semantically meaningful.

\noindent \textbf{Extensibility.}
\method is agnostic to most UMM architectures and post-training approaches.
First, since it is a general training framework that only introduces a lightweight trainable perturber, it is flexible to be integrated into most UMMs.
Second, it does not conflict with existing methods, but can serve as their complement for further improvement on consistency and performance. We further explore the intergration of \method with emerging post-training method e.g.,~\citep{jiang2025t2i,wang2024reconstruction}, we train their post-trained model and demonstrate further improvements (see \S\ref{sec-exp-general-effi}). 
%%%%%%%%%%%%%%%%%%%%%%%%%%%%%%%%%%%%%%%%%%%%%%%%%%%%%%%%%%%%%%%%% section 4

\section{Experiments}
\label{sec:experiments}

% In this section, we conducted extensive experiments to evaluate the proposed \method.
 
\subsection{Experimental Setup}

\noindent \textbf{Tasks and datasets.}
We evaluated \method on popular benchmarks.
Specifically, VQAv2 \cite{balanced_vqa_v2}, MMMU~\cite{yue2024mmmu}, POPE~\cite{li2023evaluating}, and MMBench ~\citep{liu2024mmbench} are adopted for understanding evaluation and GenEval~\cite{ghosh2023geneval} is employed for generation evaluation.
For the evaluation of consistency, we report WISE score~\citep{niu2025wise} and UnifiedBench~\cite{yan2025can}.
For OOD robustness, we adopt NaturalBench~\cite{li2024naturalbench}, a challenging benchmark comprising real-world images captured in natural, uncontrolled environments (e.g., low lighting, occlusion, unusual viewpoints) that test robust visual reasoning. 
For adversarial robustness, we adopt AdVQA~\cite{li2021adversarial}, an adversarially constructed VQA dataset where questions are intentionally designed to mislead models through linguistic ambiguity and visual distractors.

\noindent \textbf{Implementation details.}
We implemented \method on a popular Janus-Pro-7B~\citep{chen2025janus} UMM for main experiments, and further validated with two toy models that simulate distinct UMMs designs, similar to UAE \cite{yan2025can}. 
The perturber is implemented as a 3-layer MLP operating on the shared visual-token space $\mathbb{R}^{N\times H}$, adding only $2.1$M parameters and outputting token-wise perturbations followed by normalization and clipping.
We adopt the open-source VQAv2 training set and CC3M~\citep{sharma2018conceptual}.
More training details are in \append{sec-append-train}.

\definecolor{NatRed}{RGB}{238,102,119}     % 最好 - 淡红色
\definecolor{NatBlue}{RGB}{68,119,170}     % 次好 - 蓝色
\definecolor{LightRed}{RGB}{255,230,233}   % 淡红色背景
\definecolor{LightBlue}{RGB}{230,240,250}

\begin{table}[t!]
\centering
\caption{Consistency\protect\footnotemark evaluation on UnifiedBench and WISE ($\uparrow$). 
Models marked with$^\dagger$ denote our base model.}
\vspace{-.1in}
\begin{threeparttable}
\label{tab-consistency}
\resizebox{.48\textwidth}{!}{
\begin{tabular}{lccccc}
\toprule
\textbf{Model} & \textbf{Params} & \textbf{UnifiedBench} & \textbf{WISE} & \textbf{Avg} & \textbf{Consistency Score}\\
\midrule
BAGEL~\citep{deng2025emerging}  &  \cellcolor{LightRed}14B  & \cellcolor{LightBlue}$83.48$ & \cellcolor{LightBlue}$0.41$ & \cellcolor{LightBlue}$41.95$ & \cellcolor{LightBlue}$66.49$\\
UniWorld-V1~\citep{lin2025uniworld}  & \cellcolor{LightBlue}12B & $78.99$ & $0.35$ & $39.67$ & $61.39$\\ 
BLIP-3o~\citep{chen2025blip3}  & 8B & $76.56$ & $0.39$ & $38.48$ & $61.54$ \\
OmniGen2~\citep{xiao2025omnigen} & 7B & $83.31$ & $0.30$ & $41.81$ & $61.99$\\
Janus-Pro\textsuperscript{\dag}~\citep{chen2025janus} & 7B & $82.77$ & $0.35$ & $41.54$& $63.66$ \\
Harmon~\citep{wu2025harmonizing} & 1.5B & $65.41$ & $0.41$ & $32.90$ & $55.65$\\
Show-o~\citep{xie2024show} &  1.3B  & $69.16$ & $0.30$ & $34.73$ & $53.50$\\
\midrule
Janus-Pro+SFT~\citep{chen2025janus}  &   7B  & $ 83.20 $ & $ 0.37 $ & $ 41.79 $ & $64.72 ~(+1.06)$\textsuperscript{\ddag}\\
Harmon+RecA~\citep{wang2024reconstruction} & 1.5B& $66.94$ & $0.40$ & $33.67$ & $56.16~ (+0.51)$\textsuperscript{\ddag}\\
\textbf{Janus-Pro+\method} & 7B & \cellcolor{LightRed}$ \mathbf{85.20}$ & \cellcolor{LightRed} $\mathbf{0.43}$ & \cellcolor{LightRed}$\mathbf{42.82}$ & \cellcolor{LightRed}$\mathbf{68.32 ~(+4.66)}$\textsuperscript{\ddag}\\
\bottomrule
\end{tabular}
\vspace{-.1in}
}
\end{threeparttable}
\end{table}
\footnotetext{Consistency Score = $0.6 \times \text{UnifiedBench} + 0.4 \times (\text{WISE} \times 100)$, jointly assessing self-consistency in understanding generated content (UnifiedBench) 
and prompt-image alignment (WISE), the $0.6/0.4$ weighting reflects the evaluation data ratio. \textsuperscript{\ddag} show improvement over base model.}

\noindent \textbf{Baselines.}
We compare \method to two categories of baselines: 
(1) \textbf{Different UMMs:} 
(i) \textit{Auto-regressive models}~\cite{chen2025janus,team2024chameleon,qu2025tokenflow,ge2024seed,xie2024show}, which unify understanding and generation through next-token prediction in a shared token space, with improvements in vision tokenizers; 
(ii) \textit{Diffusion-based models}~\cite{chen2025blip3,xiao2025omnigen}, which leverage latent diffusion for generation while maintaining autoregressive understanding (e.g., BLIP-3o~\citep{chen2025blip3}, OmniGen2~\citep{xiao2025omnigen}); 
(iii) \textit{Hybrid architectures}~\cite{lin2025uniworld,deng2025emerging}, which employ specialized modules for different modalities (e.g., UniWorld-V1~\citep{lin2025uniworld}, BAGEL~\citep{deng2025emerging}). 
(2) \textbf{Post-training methods:}
(i) \textit{Reconstruction-based alignment}, which uses caption-then-reconstruct cycles to enhance understanding-generation consistency (e.g., RecA~\cite{wang2024reconstruction}); 
(ii) \textit{Reward-based approaches}, which use reward function to refine their outputs(e.g,~\citep{jiang2025t2i})
Unlike these methods, \method introduces decoder-constrained self-adversarial training, converting latent inconsistencies into visually coherent counterexamples to improve reasoning robustness while preserving generation fidelity.

\subsection{Consistency Evaluation}
\label{sec-exp-consist}
We evaluated consistency on two benchmarks: UnifiedBench~\citep{yan2025can} and WISE score~\citep{niu2025wise}.
UnifiedBench is a reconstruction-based benchmark tailored for UMMs, where the ``caption--generate--compare'' protocol measures how consistently information is preserved when images is converted into text and decoded back into images; the unified score is defined as the similarity between the ground truth and reconstructed images.
WISE is a world-knowledge-informed text-to-image benchmark with 1000 knowledge-intensive prompts that are scored by a multimodal judge along consistency with the prompt, realism, and aesthetic quality; its overall Score emphasizes how faithfully the generated image reflects the textual description, making it a natural testbed for text-to-image consistency.

We followed Protocol 1 from \citep{yan2025can} to evaluate the consistency of \method. 
Specifically, we randomly sampled $100$ images from LAION-5B~\citep{schuhmann2022laion} as a test set. 
For each image, the model first generated a caption (understanding), based on which the model then reconstructed the image (generation).
Finally, we computed the similarity scores between the original and reconstructed images using four vision-language backbones (CLIP \cite{radford2021learning}, SigLIP \cite{zhai2023sigmoid}, DINO-v2 \cite{oquab2023dinov2}, and DreamSim \cite{fu2023dreamsim}). 

As shown in \Cref{tab-consistency}, \method substantially improves the unification score across all metrics compared to both the Janus-Pro baseline and SFT-only training. 
This improvement demonstrates that self-play training not only enhances understanding, but also strengthens the bidirectional consistency between the understanding and generation branches, forcing the model to maintain semantic coherence across modalities and reducing the understanding-generation gap inherent in dual unified architectures.

\definecolor{NatRed}{RGB}{238,102,119}     % 最好 - 淡红色
\definecolor{NatBlue}{RGB}{68,119,170}     % 次好 - 蓝色
\definecolor{LightRed}{RGB}{255,230,233}   % 淡红色背景
\definecolor{LightBlue}{RGB}{230,240,250}

\begin{table}[t!]
  \centering
  \caption{Results for understanding benchmarks ($\uparrow$).}
  \vspace{-.1in}
  \label{Understanding_Bench}
  \setlength{\tabcolsep}{4pt}
  \renewcommand{\arraystretch}{1.00}
  \footnotesize
  \resizebox{.45\textwidth}{!}{
  \begin{tabular}{
    l
    c  % Params (B)
    S[table-format=2.2]  % VQAv2 (%)
    S[table-format=2.1]  % MMMU
    S[table-format=2.1]  % MMBench
    S[table-format=2.1]  % POPE
    S[table-format=2.1]  % overall
  }
    \toprule
    \textbf{Model}
      & {\textbf{Params}}
      & {\textbf{VQAv2$_{test}$}}
      & {\textbf{MMMU}}
      & {\textbf{MMBench}}
      & {\textbf{POPE}}
      & {\textbf{Overall}}\\
    \midrule
    TokenFlow-XL \cite{qu2025tokenflow} & \cellcolor{LightRed}14B & 77.6 & 43.2 & 76.8 & \cellcolor{LightBlue}87.8 & 71.3\\
    BAGEL \cite{deng2025emerging}             & \cellcolor{LightRed}14B  & \NA  & \cellcolor{LightBlue}55.3 & \cellcolor{LightRed}85.0 & \NA  & \NA \\
    UniWorld-V1 \cite{lin2025uniworld}        & \cellcolor{LightBlue}12B  & \NA  & \cellcolor{LightRed}58.6  & 83.5  & \NA  & \NA\\
    BLIP3-o \cite{chen2025blip3}         & 8B & \cellcolor{LightBlue}83.1 & 50.6 & \cellcolor{LightBlue}83.5 & \NA  & \NA\\
    Emu3 \cite{wang2024emu3}             & 8B  & 75.1 & 31.6 & 58.5 & 85.2 & 62.6\\
    SEED-X \cite{ge2024seed}                  & 7B  & 71.2 & 35.6 & 70.1 & 84.1 & 65.2\\
    Chameleon \cite{team2024chameleon}   & 7B & 66.0 & 22.4 & \NA  & \NA  & \NA \\
    OminiGen2 \cite{xiao2025omnigen}          & 7B  & \NA  & 53.1 & 79.1 & \NA  & \NA\\
    Liquid \cite{wu2024liquid}                & 7B  & 63.5 & \NA  & \NA  & 76.8 & \NA\\
    Janus-Pro\textsuperscript{\dag} \cite{chen2025janus}       & 7B  & 78.2 & 41.0 & 79.2 & 87.4 & 71.4\\
    Show-o \cite{xie2024show}          & 1.3B  & 69.4 & 26.7 & \NA  & 80.0 & \NA \\
    \midrule
    SFT                                      & 7B  & 79.5 & 41.2 & 79.5 & 87.6 & \cellcolor{LightBlue}71.9\\
    RecA \cite{wang2024reconstruction}        & 1.5B  & \NA  & 35.7 & \NA  & 83.9 & \NA \\
    UAE \cite{yan2025can}                     & \NA  & \NA  & \NA  & \NA  & \NA  & \NA \\
    Ours                                     & 7B  & \cellcolor{LightRed}\textbf{83.4} & 43.8 & 83.2 & \cellcolor{LightRed}\textbf{89.6} & \cellcolor{LightRed}\textbf{75.0}\\
    \bottomrule
  \end{tabular}
  }
\end{table}

\begin{table}[t!]
  \centering
  \caption{Results of text-to-image generation on GenEval~\citep{ghosh2023geneval} ($\uparrow$).}
  \label{tab:geneval}
  \vspace{-.1in}
  \setlength{\tabcolsep}{4pt}
  \footnotesize
  \resizebox{\linewidth}{!}{%
  \begin{tabular}{
    l
    S[table-format=1.2]
    S[table-format=1.2] 
    S[table-format=1.2] 
    S[table-format=1.2]
    S[table-format=1.2] 
    S[table-format=1.2] 
    S[table-format=1.2]
  }
    \toprule
    Model & {S. Obj.} & {Two Obj.} & {Counting} & {Colors} & {Position} & {Color Attri.} &\multicolumn{1}{c}{Overall}  \\
    \midrule
    SEED-X {\cite{ge2024seed}} & 0.97 & 0.58 & 0.26 & 0.80 & 0.19 & 0.14 & 0.49 \\
    Show-o \cite{xie2024show}                              & 0.95 & 0.52 & 0.49 & 0.82 & 0.11 & 0.28 & 0.53 \\
    D\,-DiT \cite{li2025dual}                             & 0.97 & 0.80 & 0.54 & 0.76 & 0.32 & 0.50 & 0.65 \\
    TokenFlow-XL \cite{qu2025tokenflow}                        & 0.95 & 0.60 & 0.41 & 0.81 & 0.16 & 0.24 & 0.55 \\
    Chameleon \cite{team2024chameleon}                           & \multicolumn{1}{c}{—} & \multicolumn{1}{c}{—} & \multicolumn{1}{c}{—} & \multicolumn{1}{c}{—} & \multicolumn{1}{c}{—} & \multicolumn{1}{c}{—} & 0.39 \\
    OminiGen2 \cite{xiao2025omnigen}                            & 0.99 & 0.86 & 0.64 & 0.85 & 0.31 & 0.55 & 0.70 \\
    Janus-Pro\textsuperscript{\dag} \cite{chen2025janus}                        & 0.99 & 0.89 & 0.59 & 0.90 & 0.79 & 0.66 & 0.80 \\
    \midrule
    SFT                        & 0.99 & 0.90 & 0.60 & 0.91 & 0.80 & 0.65 & 0.81 \\
    UAE \cite{yan2025can}                         & \multicolumn{1}{c}{—} & \multicolumn{1}{c}{—} & \multicolumn{1}{c}{—} & \multicolumn{1}{c}{—} & \multicolumn{1}{c}{—} & \multicolumn{1}{c}{—} & \cellcolor{LightRed}0.86 \\
    RecA \cite{wang2024reconstruction}                        & 1.00 & 0.98 & 0.71 & 0.93 & 0.76 & 0.77 & \cellcolor{LightRed}0.86 \\   

    Ours       & 0.99 & 0.91 & 0.62 & 0.93 & 0.80 & 0.68 & \cellcolor{LightBlue}0.82 \\
    \bottomrule
  \end{tabular}
  }
\end{table}

\subsection{Benchmark Results}
\label{sec-exp-main}

\noindent \textbf{\method improves both understanding and generation.}
\Cref{Understanding_Bench} and \ref{tab:geneval} show the results on understanding and generation benchmarks, respectively.
The results demonstrate significant improvements of \method over most competitors in both tasks.
Specifically for understanding tasks, \method shows an average improvement of $3.1\%$ over SFT, and $3.6\%$ over the baseline model.
\method also outperforms other larger models such as TokenFlow-XL, Emu3, and BLIP-3o, demonstrating that creating adversarial examples can serve as an effective augmentation approach.
As for generation, \method outperforms most competitors such as TokenFlow-XL and OminiGen, achieving stronger performance than the base model and SFT.
Note that \method performs slightly worse than UAE and RecA ($0.82$ vs. $0.86$), which is mainly due to UAE and RecA's explicit post-training on generation tasks, while ours was primarily trained on understanding tasks.

\begin{figure*}[t!]
  \centering
  \begin{subfigure}{\linewidth}
    \centering
    \includegraphics[width=\linewidth]{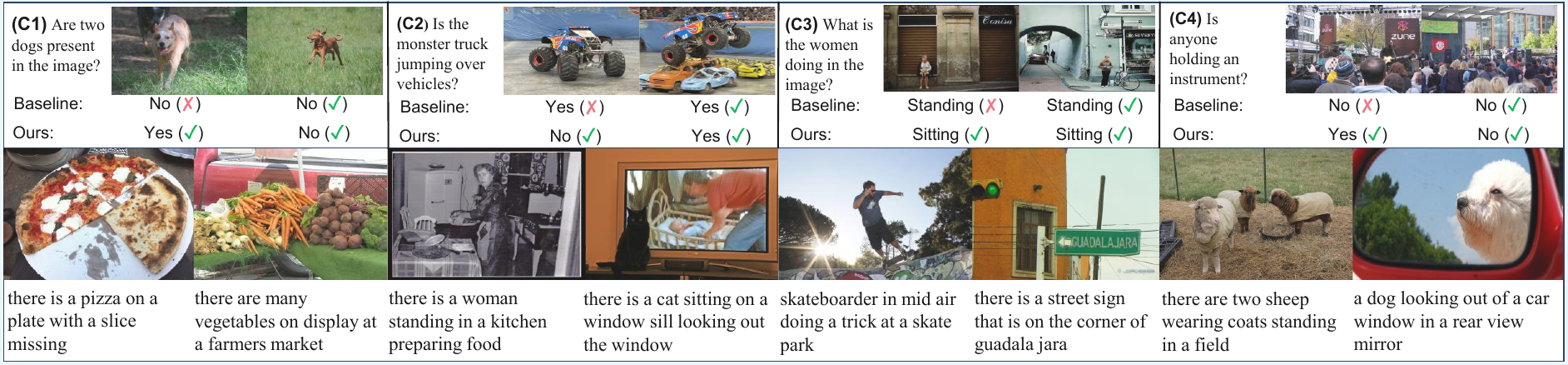}
    \caption{Case study for close-ended and open-ended understanding tasks.}
    \label{fig:casestudy}
  \end{subfigure}
  \vspace{2pt} 
  \begin{subfigure}{\linewidth}
    \centering
    \includegraphics[width=\linewidth]{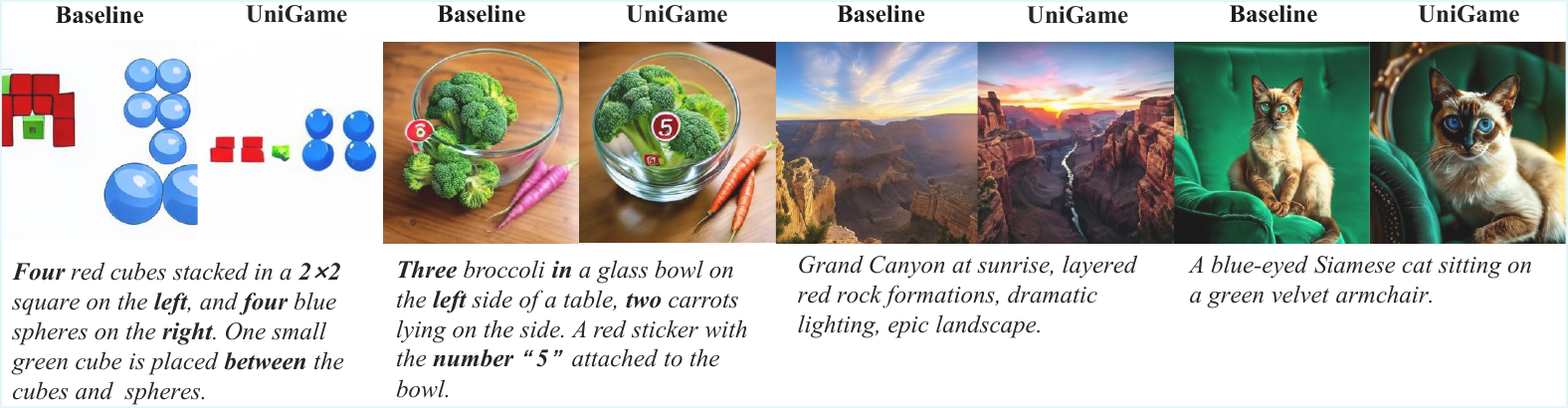}
    \caption{Case study for generation tasks.}
    \label{fig:case_study_gen}
  \end{subfigure}
  \vspace{-.3in}
  \caption{Qualitative case studies of \method understanding and generation.}
  \label{fig:case_study_all}
  \vspace{-.2in}
\end{figure*}

\noindent \textbf{\method improves model robustness.}
We evaluated the OOD and adversarial robustness on NaturalBench~\cite{li2024naturalbench} and AdVQA \cite{li2021adversarial}, respectively.
Following~\citet{li2024naturalbench}, we report \textit{Group Accuracy (G-Acc)}, 
which awards one point only when a model correctly answers all four 
(image, question) pairs in a test sample.
For AdVQA, we report standard accuracy as used in prior work. 
As shown in \Cref{fig:robustness}, \method exhibits significant improvement in OOD and adversarial benchmarks (\SI{4.8}{\percent} and \SI{6.2}{\percent} gain, respectively), suggesting its strong performance in robustness, confirming \method effectively expand the decision boundaries, as in \Cref{fig:casestudy} we explicitly probe four fine-grind visual reasoning cases, where base models fail but \method reasoned correctly.
For space reasons, we present full experiments, additional ablations, and detailed OOD/adversarial robustness breakdowns in \append{sec_app-experiment_details} and \ref{sec-app-results-robustness}.

\definecolor{NatBlue}{RGB}{68,119,170}   
\definecolor{NatRed}{RGB}{238,102,119}   
\definecolor{NatGreen}{RGB}{34,136,51}   
\definecolor{NatGold}{RGB}{204,187,68}   
\definecolor{NatPurple}{RGB}{170,51,119} 
\definecolor{NatCyan}{RGB}{102,204,238}  

\begin{figure}[t!]
  \centering
  \begin{subfigure}[b]{0.48\columnwidth}
    \centering
    \begin{tikzpicture}
      \begin{axis}[
        ybar,
        single ybar legend,
        bar width=3pt,
        width=\linewidth,
        height=3.2cm,
        ylabel={Score},
        ylabel style={font=\fontsize{8}{9.5}\selectfont},  
        symbolic x coords={Base Model, +SFT, +Ours},
        xtick=data,
        x tick label style={font=\fontsize{6}{7}\selectfont},  
        y tick label style={font=\fontsize{6}{7}\selectfont},
        legend style={
          at={(0.5,1.05)}, 
          anchor=south, 
          legend columns=2, 
          font=\fontsize{6.5}{7}\selectfont,
          draw=none,
          fill=none,
        }, 
        ymin=10, ymax=35,
        enlarge x limits=0.22,
      ]
        \addplot+[fill=NatBlue!80, draw=NatBlue] coordinates {(Base Model,23.6) (+SFT,24.5) (+Ours,28.4)};
        \addplot+[fill=NatRed!25, draw=NatRed]   coordinates {(Base Model,21.1) (+SFT,22.2) (+Ours,27.8)};
        \legend{NaturalBench, AdVQA}
      \end{axis}
    \end{tikzpicture}
    \caption{OOD and adv. robustness.}
    \label{fig:robustness}
  \end{subfigure}
  \hfill
  \begin{subfigure}[b]{0.49\columnwidth}
    \centering
    \includegraphics[width=\textwidth]{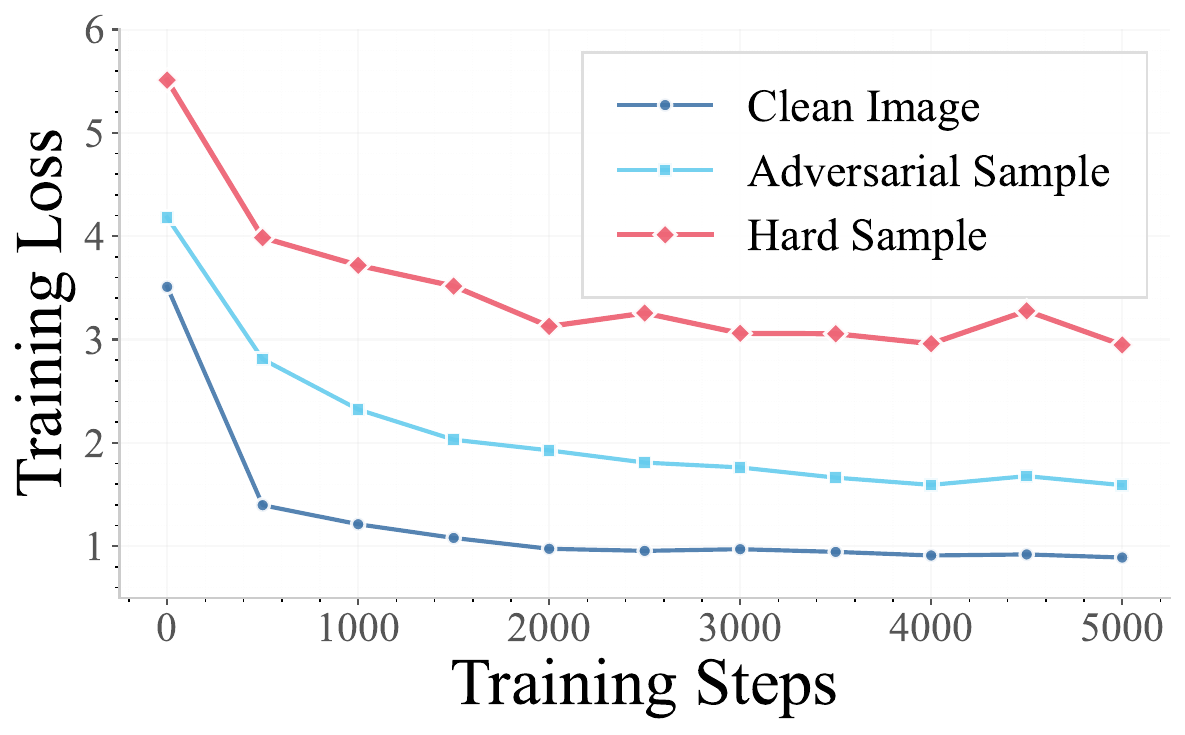}
    \caption{Training Loss Comparison}
    \label{fig:hard_loss_compare}
  \end{subfigure}
  \vspace{-.1in}
  \caption{(a) Robustness evaluation. (b) We observe that over 5K of training steps, the hard-sample loss persistently dominates that of Clean/Adversarial, suggesting \method continuously generates samples that are most challenging for the current model state.}
  \label{fig:two_piller}
  \vspace{-.1in}
\end{figure}
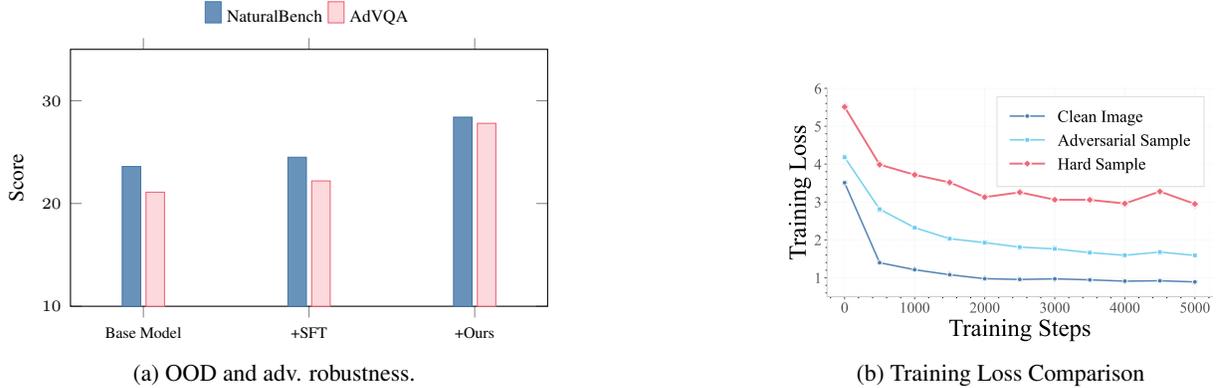

\begin{table}[t]
\centering
\caption{\textbf{Ablation: UniGame vs embedding perturbation.} All methods use the same perturbation budget ($\varepsilon_{\mathrm{max}}{=}0.02$), update schedule, and training steps (16k). For embedding-only baselines, we apply perturbations directly in the visual token space without decoding. We report VQAv2 accuracy (\%).}
\vspace{-.1in}
\label{tab:vs_embed}
\setlength{\tabcolsep}{6pt}
\renewcommand{\arraystretch}{1.00}
\footnotesize
\resizebox{0.6\linewidth}{!}{%
\begin{tabular}{@{}lc@{}}
\toprule
\textbf{Method} & \textbf{Performance} \\
\midrule
\rowcolor{gray!10}
Baseline (SFT) & $79.5$ \\
\midrule
\multicolumn{2}{@{}l}{\textbf{\textit{Embedding-only perturbation}}} \\
\quad Random noise in token space & $78.5$ \\
\quad Adversarial emb.  & $78.9$ \\
\quad Adv. emb. + Cosine Similarity  & $79.6$ \\
\quad Adv. emb. + Cosine + Buffer & $80.2$ \\
\midrule
\multicolumn{2}{@{}l}{\textbf{\textit{Decoder-constrained perturbation (Ours)}}} \\
\quad Decoding only   & $81.5$ \\
\quad Decoding + Cosine Similarity & $82.2$ \\
\quad Decoding + CLIP & \cellcolor{LightBlue}$82.7$ \\
\quad Full (+ CLIP + Buffer) & \cellcolor{LightRed}$\mathbf{83.4}$ \\
\bottomrule
\end{tabular}
}
\vspace{-.2in}
\end{table}

\subsection{Ablation Study}
\label{sec-exp-ablation}
\Cref{tab:vs_embed} shows the comparison between the embedding-only and decoder-constrained adversarial perturbations under matched settings. 
For embedding-only baselines, we apply perturbations directly in the visual token space without decoding, using cosine similarity constraints to prevent excessive token drift. 
The strongest embedding baseline incorporating adversarial perturbations, token-space cosine constraints, and buffer replay achieves $80.2\%$ accuracy on VQAv2, representing a modest $+0.7\%$ improvement over the SFT baseline.
In contrast, our decoder-constrained approach forces perturbations to pass through the model's native decoder, rendering adversarial tokens into realistic images before evaluation. 
Notably, even \emph{without} CLIP filtering, decoding alone improves accuracy to $81.5\%$ ($+2.0\%$ over SFT and $+1.3\%$ over embedding perturbation), demonstrating that on-manifold constraints are inherently superior to token-space constraints. 
When we apply cosine similarity constraints in the decoded image feature space, performance further increases to $82.2\%$.
Replacing feature-level cosine with CLIP's text-image semantic matching yields $82.7\%$, validating that semantic constraints outperform purely geometric ones.

We further ablate the perturber and hard-sample buffer capacity.
For the perturber, a 3-layer MLP (83.4\%) outperforms both the 2-layer variant (82.8\%) and the deeper 4-layer variant (81.2\%), indicating that moderate capacity best balances expressiveness.
For the buffer, a size of 50 yields the best accuracy (83.4\%), while smaller sizes of 30 (83.1\%) and 10 (82.5\%) provide insufficient diversity.
Key insights: 
(i) the embedding-level perturbations can only leverage weak adversarial signals ($+0.7\%$) because they operate in an abstract space disconnected from visual semantics; 
(ii) decoder constraints enforce on-manifold perturbations, yielding stronger adversarial training ($+2.0\%$); 
(iii) using CLIP to maintain semantics further amplifies gains by ensuring adversarial samples remain semantically consistent with the query text.
Together, these components establish a principled framework for self-play training in UMMs. 

% \paragraph{Adversarial budget tracking.}
% We log the learned $\varepsilon$ at each step and plot its trajectory (Fig.~\Cref{fig:eps-curve}); values remain safely bounded under $\varepsilon_{\max}{=}0.02$.

\subsection{Case Study}
\label{sec-exp-case}
We provide case studies on both understanding and generation tasks for qualitative analysis. 

\noindent \textbf{Understanding tasks.}
\Cref{fig:casestudy} illustrates four challenging categories of visual reasoning tasks: object counting, object interaction, spatial relation and location, and crowd object detection.
\method outperformed the baseline models in all scenarios.
For instance, in C4 (crowd object detection), dense and overlapping objects in crowded scenes challenge both localization and recognition. 
The baseline produces vague or incorrect answers, whereas \method maintains accuracy by learning from decoded adversarial samples that emphasize occlusion and clutter.
These improvements align with our quantitative gains, confirming that \method systematically addresses decision-critical reasoning failures rather than merely fitting to benchmark statistics. And we evaluate the open-ended captioning task in \Cref{fig:casestudy}, the qualitative examples align with our quantitative gains on benchmarks, and suggest that \method helps the model move toward semantically richer and accurate descriptions.
More analysis is in \append{sec-append-case-under}.

\noindent \textbf{Generation tasks.}
\Cref{fig:case_study_gen} compares generations from the same prompts before and after post-training with \method.
Overall speaking, \method helps UMMs to generate more faithful, accurate, and stylistic images.
For instance, on the synthetic shapes example, the baseline model already produces plausible objects but often violates fine-grained layout constraints (e.g., incorrect left/right ordering or cube–sphere counts), whereas \method yields images that respect the specified $2\times 2$ red cube stack, the correct number of blue spheres, and the spatial relations such as “on the left / on the right” and “between”.
More explanations of other cases are in \append{sec-append-case-gen}.
Together with the understanding cases, it suggests that \method enhances cross-modal consistency without sacrificing, and in some cases even improving the generation quality.

% \subsection{Theoretical View}
% \label{sec:theory}

% \textbf{Distributional view.}
% Let $\mathcal{D}_{\mathrm{clean}}$ be the empirical training distribution.
% \method adds decoded, semantically filtered perturbations, inducing an effective training distribution
% \[
% \mathcal{D}_{\mathrm{mix}}
% = (1-\mu)\,\mathcal{D}_{\mathrm{clean}} + \mu\,\mathcal{D}_{\mathrm{pert}},
% \]
% where $\mathcal{D}_{\mathrm{pert}}$ collects on-manifold variants of fragile cases and $\mu\in(0,1)$ is the sampling ratio.
% Thus the model is trained on a more diverse but still realistic neighborhood of the original data, with extra mass concentrated on decision-critical regions (e.g., counting, fine attributes, occlusions).

% \textbf{Optimization view.}
% Practically, \method minimizes a single coupled objective over $(\theta_U,\theta_C)$ and uses alternating updates (first $\theta_C$, then $\theta_U$), i.e., block-coordinate stochastic gradient descent.
% Under standard smoothness and bounded-variance assumptions, such schemes are known to converge to first-order stationary points of the joint objective, which is consistent with the stable co-adaptation we observe in practice.

\subsection{Extensibility and Efficiency}
\label{sec-exp-general-effi}

\method remains agnostic to UMM architectures and is computationally efficient compared to other post-training methods.
In this section, we evaluate its generality and efficiency using the full set of VQAv2 on 2×H100 (80 GB) with mixed precision.
Unless noted otherwise, we use image generation size $384$ and a global batch size of $8$.

\begin{table}[t]
  \centering
  \caption{\method can be plugged into an existing post-training pipeline with modest extra training to jointly improve understanding, generation, and unification. Starting from a RecA-trained model harmon $1.5$B, we further train with $5$K UniGame steps ($\sim 10$ GPU-h), yielding consistent improvements.}
  \label{tab:after_reca}
  \vspace{-.1in}
  \setlength{\tabcolsep}{4pt}
  \renewcommand{\arraystretch}{1.05}
  \footnotesize
  \resizebox{.38\textwidth}{!}{
  \begin{tabular}{
    l
    c      % MMMU
    c      % GenEval
    c      % UnifiedBench
  }
    \toprule
    \textbf{Method}
      & \multicolumn{1}{c}{\textbf{MMMU}}
      & \multicolumn{1}{c}{\textbf{GenEval}} 
      & \multicolumn{1}{c}{\textbf{UnifiedBench}}\\
    &
      \multicolumn{1}{c}{\emph{understanding}}
      & \multicolumn{1}{c}{\emph{generation}}
      & \multicolumn{1}{c}{\emph{ocnsistency}}\\
    \midrule
    RecA
      & $35.7$    
      & $0.86$    
      & $66.94$     
      \\
    RecA + Ours
      & \cellcolor{LightRed}$36.2$ \textcolor{NatRed}{(\textbf{$+0.5$})}
      & \cellcolor{LightRed}$0.86$  \textcolor{NatRed}{(\textbf{$--$})}
      & \cellcolor{LightRed}$68.21$ \textcolor{NatRed}{(\textbf{$+1.27$})}
      \\
    \bottomrule
  \end{tabular}
  }
  \vspace{-.1in}
\end{table}

\noindent \textbf{Extensibility.}
We implement \method as a complement to RecA~\citep{wang2024reconstruction}.
\Cref{tab:after_reca} shows consistent performance gains. method outperforms the RecA by $0.5$ on MMMU for understanding and $1.27$ on UnifiedBench for consistency, while remaining the same on GenEval.
These results indicate that \method can serve as a lightweight, plug-and-play post-training module that can be integrated into existing pipelines, requiring only minimal additional computation.

In addition to RecA, we further constructed two architectures: (1) \textit{UMM-1} uses a Qwen2.5-VL~\citep{yang2025qwen3} backbone with a SigLIP2~\citep{zhai2023sigmoid} understanding encoder and a Stable Diffusion-1.5~\citep{Rombach_2022_CVPR} image branch;
(2) \textit{UMM-2} keeps the vision/generation stack unchanged and replaces the backbone with GPT-OSS~\citep{openai2025gptoss120bgptoss20bmodel}. 
Since GPT-OSS is designed for texts, we inserted a trainable 2-layer MLP that projects vision embeddings into the language space.
We further applied \method to 
two backbones: 
BLIP-3o~\citep{chen2025blip3} and 
Chameleon~\citep{team2024chameleon}. All show consistent gains (\append{sec-append-backbone}).
\Cref{tab:toymodel} shows that \method is agnostic to model architectures and can improve the performance of different backbones.
Moreover, the gains are achieved with fewer trainable parameters (e.g., $\sim$0.45\% for UMM-2 and $\sim$1.43\% for UMM-1), indicating its parameter-efficient generalization.
% To decouple algorithmic effects from scale and training recipe, we conduct all main analyses under a general, modular toy setting that fixes the shared visual-token interface while swapping only the language backbone.

\definecolor{NatBlue}{RGB}{68,119,170}   
\definecolor{NatRed}{RGB}{238,102,119}   
\definecolor{NatGreen}{RGB}{34,136,51}   
\definecolor{NatGold}{RGB}{204,187,68}   
\definecolor{NatPurple}{RGB}{170,51,119} 
\definecolor{NatCyan}{RGB}{102,204,238}  

\begin{table}[t!]
  \centering
  \caption{Extensibility analysis using two toy backbones.}
  \label{tab:toymodel}
  \vspace{-.1in}
  \footnotesize
  \resizebox{.48\textwidth}{!}{
  \begin{tabular}{lccc}
    \toprule
    Model  & Baseline & +\method & Trainable\\
    \midrule
    UMM-1 (Qwen2.5-VL) 
      & $60.4$ 
      & $66.4\,\color{NatRed}{~(+6.0\%)}$ 
      & $\sim\,1.43\%~(100.3\mathrm{M}/7\mathrm{B})$ \\
    UMM-2 (GPT-OSS)  
      & $28.9$   
      & $53.2\,\color{NatRed}{~(+24.3\%)}$ 
      & $\sim\,0.45\%~(133.9\mathrm{M}/30\mathrm{B})$\\
    \bottomrule
  \end{tabular}
  }
\end{table}

% \Cref{tab:toymodel} compares \method on two toy backbones: (i) \textbf{Toy-1} with an end-to-end VL–pretrained backbone (Qwen2.5-VL), and (ii) \textbf{Toy-2} with a \emph{pure-language} backbone (GPT-OSS) plus a light vision projector.
% Despite lacking cross-modal priors, the pure-language \textbf{Toy-2} closes much of the gap after applying \method, reaching \textbf{Toy-1}'s zero-shot accuracy after $\sim$50k \method steps.
% This suggests that \method can \emph{induce} vision–language alignment and compositional priors in models without VL pretraining capabilities previously associated with large-scale pipelines such as LLaVA~\citep{liu2023visual}.

% The results in \Cref{tab:toymodel} show that, compared to the end-to-end VL pretrained backbone, the pure-language OSS backbone achieved Toy-1's zero-shot accuracy after $\sim$50k UniGame steps.
% Although this result thanks to the scale of OSS model, but still surprisingly shows UniGame can generalize to a model that lacks a cross-modal latent knowledge, effectively inducing vision-language alignment and priors, which was previously only achievable through large-scale pre-training like llava~\citep{liu2023visual}.

\noindent \textbf{Efficiency.}
We further evaluate the efficiency of \method in comparison with RecA~\citep{wang2024reconstruction} and UAE~\citep{yan2025can} on MMMU. \Cref{tab:train-time} shows that while achieving stronger performance, \method uses fewer trainable parameters, indicating its efficiency over existing post-training approaches.

\definecolor{NatBlue}{RGB}{68,119,170}   
\definecolor{NatRed}{RGB}{238,102,119}   
\definecolor{NatGreen}{RGB}{34,136,51}   
\definecolor{NatGold}{RGB}{204,187,68}   
\definecolor{NatPurple}{RGB}{170,51,119} 
\definecolor{NatCyan}{RGB}{102,204,238}  

\begin{table}[t!]
  \centering
  \footnotesize
  \caption{Efficiency study. Trainable parameter ratios are estimated from the official repository.}
  \label{tab:train-time}
  \vspace{-.1in}
  \resizebox{.38\textwidth}{!}{
  \begin{tabular}{lccc}
    \toprule
    Method  & Baseline & +\method & Trainable \\
    \midrule
    ReCA    
      & $34.7$  
      & $35.7\,\color{NatRed}{~(+1.0\%)}$ 
      & $\sim\,91\%~(\sim\,1.4\mathrm{B}/1.5\mathrm{B})$ \\
    UAE     
      & \NA   
      & \NA 
      & $\sim\,1\%~(0.1\mathrm{B}/11\mathrm{B})$ \\
    UniGame 
      & $41.0$ 
      & $43.8\,\color{NatRed}{~(+2.8\%)}$ 
      & $\sim\,1\%~(100.3\mathrm{M}/7\mathrm{B})$ \\
    \bottomrule
  \end{tabular}
  }
  \vspace{-.1in}
\end{table}

\subsection{Convergence and Hyperparameter Analysis}
\label{sec-exp-conv}
Finally, we present a large-scale analysis on the convergence and hyperparameters (e.g., the hard buffer threshold $\tau$, trade-off $\beta$, and perturbation budget $\delta$; see \Cref{fig:eps_curve}).
To study the training dynamic, we also conduct extensive ablation study on the learning rates of \textbf{\emph{two major minmax opponents}} in \Cref{fig:lr_ablation}. 
Further, we systematically study the minimax dynamics by visualizing the optimization trajectory of each run. The best configuration yields a well-behaved minimax trajectory, where the two players alternate smoothly without divergence, see \append{fig:opt-path}.
We also probe the self-play dynamics between the two opponents and clearly observe the interaction: the two branches alternately dominate the training objective, exhibiting a stable tug-of-war behavior, and change of dominance, see \Cref{fig:dynamic} and \ref{fig:dominate-timeline}.
More details in \append{sec-append-conv} demonstrate that \method offers a steady training process and stays relatively robust to different hyperparameter choices.
\append{sec:theory} further presents some theoretical insights.

% \textbf{Analysis of the Self-play Training Dynamics.}
% We ablate different design choices of UniGame see in Appendix \Cref{sec_app-experiment_details}. 、

% We vary the Generation and understanding update ratio in  $\{1{:}1,\ 1{:}5,\ 1{:}10\}$.
% The $1{:}1$ alternation gives the best clean/robust trade-off; larger Perturbers raise early difficulty but can erode clean accuracy if over-parameterized. 
% By extending Generation updates for a fixed Understanding, we observe ASR saturation followed by Understanding catch-up see Appendix ~\Cref{fig:dominate-timeline}.
% Short, interleaved schedules are therefore preferable to long Gen-generation-only phases.

\section{Conclusion and Limitation}
\label{sec:conclusion}

\method is the first self-adversarial post-training framework to improve the consistency of UMMs.
It formulated a minimax optimization game of the understanding and generation branches, thus enabling the model to autonomously discover its own failures. 
\method consistently showed increased consistency, performance, and robustness, highlighting the great potential of optimizing UMMs within the models for further improvement.

\textbf{Limitations.}
This work has following limitations.
First, we primarily evaluate Janus-Pro-7B; broader model coverage may reveal additional insights.
Second, we use a limited set of datasets, and future work should test \method on more diverse and challenging benchmarks.

\section*{Acknowledgments}
This paper is partially supported by unrestricted gift from Google, William \& Mary Faculty Research Award, and Modal Academic Compute Award.
The authors acknowledge William \& Mary Research Computing for providing computational resources and/or technical support that have contributed to the results reported within this paper. URL: https://www.wm.edu/it/rc.

{
    \small
    \bibliographystyle{ieeenat_fullname}
    \bibliography{main}
}

\newpage
\appendix
\clearpage
\setcounter{page}{1}
\maketitlesupplementary

\section*{Appendix Contents}
\label{sec-append-contents}
\begingroup
\setlength{\parindent}{0pt}   % 每行不缩进
\setlength{\parskip}{0.25em}  % 行间距稍微拉开一点

\ref{sec-append-algo}\quad Algorithm Details \dotfill \pageref{sec-append-algo}\par
\ref{sec-append-train}\quad Training Details \dotfill \pageref{sec-append-train}\par
\ref{sec_app-experiment_details}\quad Detailed Experimental Results \dotfill \pageref{sec_app-experiment_details}\par
\ref{sec-app-results-robustness}\quad Robustness Results \dotfill \pageref{sec-app-results-robustness}\par
\ref{sec-append-case}\quad Details on Case Study \dotfill \pageref{sec-append-case}\par
\ref{sec-append-conv}\quad Convergence and Hyperparameter Analysis \dotfill \pageref{sec-append-conv}\par
\ref{sec:theory}\quad Theoretical Insights \dotfill \pageref{sec:theory}\par

\endgroup

\section{Algorithm Details}
\label{sec-append-algo}

The complete training algorithm of \method is shown in \Cref{alg:unigame}.

\begin{algorithm}[htbp]
  \caption{\method}
  \label{alg:unigame}
  \begin{algorithmic}[1]
    \STATE Initialize $\theta_U$ (understanding) and $\theta_C$ (Perturber);
    \FOR{each training step $t = 1,2,\dots$}
      \STATE Sample minibatch $\{(\mathbf{x}_i, q_i, a_i)\}_{i=1}^M \sim \mathcal{D}$ and encode
             $\mathbf{z}_i = \mathrm{Proj}(\mathrm{Enc}(\mathbf{x}_i))$
      \STATE \textbf{Challenge step (update $C$):}
      \STATE Compute perturbations $\boldsymbol{\delta}_i = C(\mathbf{z}_i; \theta_C)$ and
             perturbed tokens $\tilde{\mathbf{z}}_i = \mathbf{z}_i + \boldsymbol{\delta}_i$
             with $\|\boldsymbol{\delta}_i\| \le \varepsilon_{\max}$
      \STATE Decode candidates $\tilde{\mathbf{x}}_i = G(\tilde{\mathbf{z}}_i)$
      \STATE Compute $\mathcal{L}_C(\theta_C;\theta_U)$ as in Eq.~\eqref{eq:lossC}
      \STATE Update $\theta_C \leftarrow \theta_C + \eta_C \nabla_{\theta_C}\mathcal{L}_C$
      \IF{$t \bmod m = 0$}
        \STATE Compute scores $H_j$ and keep candidates passing CLIP threshold $\tau$ and push hard examples into $\mathcal{B}$ via Eq.~\eqref{eq:buffer}
      \ENDIF
      \STATE \textbf{Understand step (update $U$):}
      \STATE Construct mixed batch: clean samples $(\mathbf{z}_i, q_i, a_i)$, and hard samples $(\hat{\mathbf{z}}_j, \hat{q}_j, \hat{a}_j)$ drawn from $\mathcal{B}$
      \STATE Compute $\mathcal{L}_U(\theta_U)$ on the mixed batch as in Eq.~\eqref{eq:lossU}
      \STATE Update $\theta_U \leftarrow \theta_U - \eta_U \nabla_{\theta_U}\mathcal{L}_U$
    \ENDFOR
  \end{algorithmic}
\end{algorithm}

\section{Training Details}
\label{sec-append-train}

\subsection{Training and Testing Data}
\label{sec-append-train-data}

\textbf{Data volumes.}
Unless otherwise noted, we follow the official training/evaluation splits and report results on the standard benchmarks. 
Training uses
\textbf{VQAV2 train-split}~\cite{balanced_vqa_v2} is a large-scale visual question answering benchmark (hundreds of thousands of image–question pairs) collected from MS-COCO images with crowd-sourced free-form answers; it emphasizes grounded visual reasoning under natural images.
\textbf{CC3M}~\cite{sharma2018conceptual} (training only) is a large web-scale image–caption corpus ($\sim$3M pairs in the full set); we use a filtered \emph{subset} of 100k as text–image supervision for the generative branch.

\noindent \textbf{Benchmarks.}
We briefly introduce the benchmarks:
\begin{itemize}
    \item \textbf{VQAv2 test-dev}~\cite{balanced_vqa_v2}: the official VQAv2 test-dev split contains \textbf{\num{104000}} questions; evaluation is via the online server.\footnote{Counts from the official VQA site; see also recent reports confirming 104K for test-dev.}
\item \textbf{MMMU}~\cite{yue2024mmmu}: a college-level, multi-discipline benchmark with \textbf{\num{11500}} questions in total (we report on the official test set).
\item \textbf{POPE}~\cite{li2023evaluating}: object-hallucination evaluation with a balanced, image-grounded design; the \textbf{test split has \num{9000}} QA pairs.
\item \textbf{MMBench}~\cite{liu2024mmbench}: curated multiple-choice suite; \textbf{dev \num{1164}} and \textbf{test \num{1784}} questions (4:6 split of $\sim$3K).
\item \textbf{GenEval}~\cite{ghosh2023geneval}: object/layout/attribute–focused T2I evaluation with \textbf{\num{553}} prompts (reference-free automatic checks).
\item \textbf{UnifiedBench}~\cite{yan2025can}: unification score via caption$\rightarrow$reconstruction; Protocol-1 uses \textbf{100 source images}.
\item \textbf{WISE}~\cite{niu2025wise}: knowledge-informed T2I evaluation with \textbf{\num{1000}} structured prompts across 25 subdomains. 
\item \textbf{NaturalBench}~\cite{li2024naturalbench}: vision-centric VQA with natural adversarial samples, $\sim$\num{10000} human-verified image–question pairs (\emph{\num{2500} groups} under the 2-image$\times$2-question protocol), scored by G-Acc.
\item \textbf{AdVQA}~\cite{li2021adversarial}: human-in-the-loop adversarial VQA; total size reported as \textbf{$\sim$\num{46807}} examples (commonly used splits include \textbf{$\sim$\num{5123}} val / \textbf{$\sim$\num{23399}} test). 
\end{itemize}

\subsection{Hyperparameter Details}
\label{sec-append-train-hyper}

\noindent \textbf{Optimization details.}
\method is like the current UMMs post-training, is an end-to-end method and involves decoding images in each batch, to balance performance and cost. Our optimizations are as followed.
We use AdamW optimizers with learning rates for Generation (\verb|gen_lr|) and Understanding (\verb|und_lr|). 
We conduct extensive ablation on the learning rate ratio between these two components (detailed in \append{sec_app-experiment_details} and Table~\ref{gd-ratio}), ultimately finding that a ratio of approximately 250 achieves optimal performance (\verb|gen_lr|$=5{\times}10^{-3}$, 
\verb|und_lr|$=2{\times}10^{-5}$).

We implement mixed precision for training, given that Uni-Game only learned and uses small-norm perturbation, insufficient numerical precision can quantize away the perturbation's gradients and wash out all the supervision. 
We vary the Generation and understanding update ratio in  $\{1{:}1,\ 1{:}5,\ 1{:}10\}$.
We performed a precision ablation comparing \texttt{fp16-all}, \texttt{bf16-all}, \texttt{tf32-enabled}, \texttt{fp32-all}, \texttt{fp16(G)+fp32(loss)}, and \texttt{bf16(G/D)+fp32(loss)}, and found that our final choice—computing the perturbation update, regularizer, and losses in \texttt{float32} while running the remaining forward/backward in \texttt{bfloat16}—consistently achieved the best stability–efficiency trade-off and the highest robust accuracy.
We force all computations that determine the perturbation and its supervision to \texttt{float32}. 
Gradient norms and per-role clipping are also applied in FP32, and optimizer states remain FP32 (AdamW default). 
All other forward/backward passes (vision tower, diffusion decoder, and LLM blocks) run under \verb|bfloat16| \verb|autocast| for throughput. 
This preserves the perturbation signal while retaining the speed benefits of mixed precision.

\subsection{Perturber}
\label{perturber}
\paragraph{Network architecture of $C$.}
We implement the perturber $C$ as a lightweight three-layer MLP that operates on each fused visual token after the language model. The first two layers have the same width as the UMM hidden size and apply non-linear transformations that refine the token representation and extract a direction in the shared visual-token space. The third layer acts as a direction head, mapping the hidden representation back to the token space and indicating along which semantic direction each token should be pushed to maximally challenge the understanding branch. In parallel, $C$ maintains a single learnable scalar gate $\varepsilon$, shared across tokens and constrained within the perturbation budget $[0,\varepsilon_{\max}]$, which controls the overall perturbation strength. In this way, one part of $C$ is responsible for discovering semantically adversarial directions, while the scalar gate $\varepsilon$ controls how strongly these directions are applied, keeping the module compact (with $|\theta_C| \ll \min(|\theta_U|,|\theta_G|)$) yet able to generate small but semantically meaningful adversarial perturbations.

\subsection{Hard Samples}
\label{sec-append-train-hard}

\begin{figure}[htbp]  
  \centering
  \includegraphics[width=\linewidth]{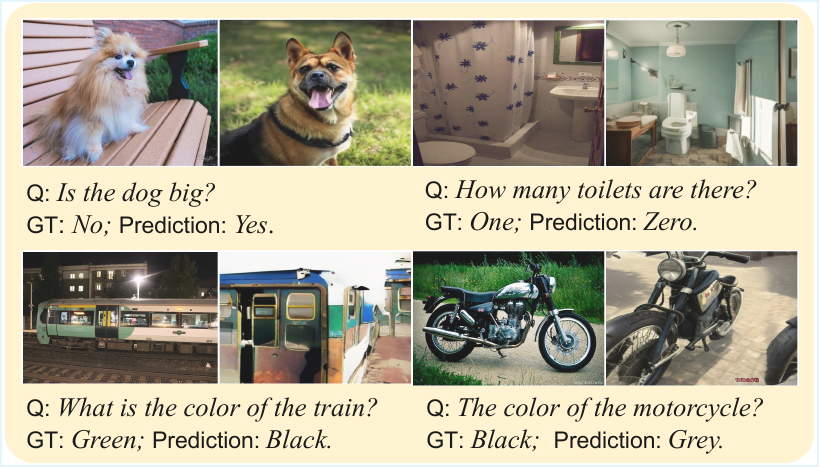}
  \caption{Cases are drawn from the hard-sample buffer and represent failure cases that successfully challenged the model.}
  \label{fig:hard_case_study}
\end{figure}

\method added a hard sampler buffer to select only the challenging adversarial samples for training.
\Cref{fig:hard_case_study} shows some challenging examples in our experiments.

\section{Detailed Experimental Results}
\label{sec_app-experiment_details}

\subsection{Additional Backbone Validation}
\label{sec-append-backbone}
To validate the generality of \method beyond Janus-Pro-7B, we applied it to two additional UMM backbones: 
BLIP-3o~\citep{chen2025blip3} (diffusion-based) and 
Chameleon~\citep{team2024chameleon} (auto-regressive), 
both trained on a 0.4 split of the VQAv2 training set 
under-matched settings.
All backbones show positive consistency gains: 
BLIP-3o (+2.5, MMMU 51.2, GenEval 0.54), 
and Chameleon (+2.2, VQAv2 40.2 with +1.7 gain, MMMU 24.0, GenEval 0.40). 
These results confirm that \method generalizes across 
different UMM architectures, including both auto-regressive and diffusion-based designs.

\subsection{Learning Rate Ratio Ablation}
\begin{figure}[t]
    \centering
    \includegraphics[width=.35\textwidth]{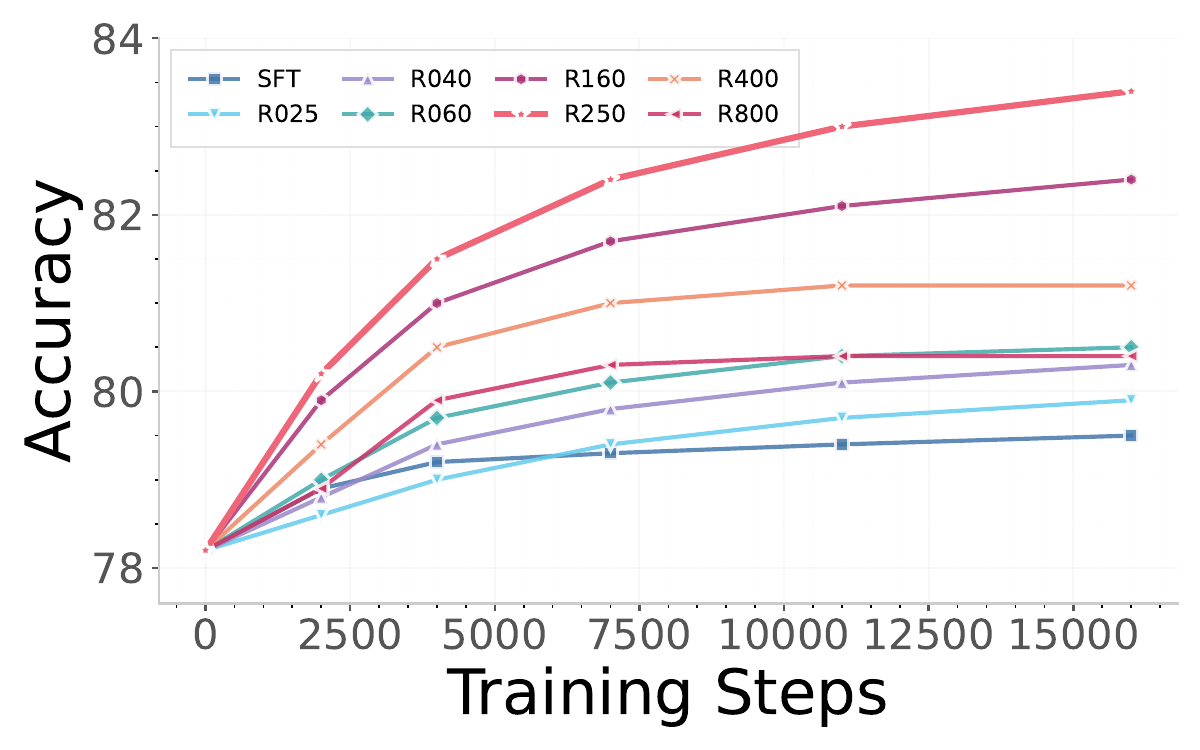}
    \caption{Training dynamics analysis. VQA accuracy evolution across different adversarial ratios, with best achieving optimal performance at 83.4\%.}
    \label{fig:lr_ablation}
\end{figure}
To determine the optimal balance between the generation and understanding branches, we conduct an extensive sweep of learning rate ratios. \Cref{gd-ratio} lists the complete set of configurations tested. 

\begin{table}[t]
  \centering
  \caption{Learning-rate configurations for the adversarial ratio sweep.
    Each row (ID R\textit{xxx}) specifies a pair of learning rates for the
    generation (\texttt{gen\_lr}) and understanding module
    (\texttt{und\_lr}); the last column reports their ratio $Gen/Und$.
    For example, R250 corresponds to \texttt{gen\_lr}$=5\times10^{-3}$ and
    \texttt{und\_lr}$=2\times10^{-5}$, i.e., a $250{:}1$ ratio.
    These IDs (R025--R800) are used in Fig.~\ref{fig:lr_ablation}(b) to plot
    validation performance as a function of the adversarial ratio.}
  \label{gd-ratio}
  \setlength{\tabcolsep}{6pt}
  \small
  \begin{tabular}{l
                  S[table-format=1.1e-1]
                  S[table-format=1.1e-1]
                  c}
    \toprule
    ID & {gen\_lr} & {und\_lr} & $Gen/Und$ \\
    \midrule
    R025 & 1.6e-3 & 6.3e-5 & $\approx 25.4$ \\
    R040 & 2e-3   & 5e-5   & 40 \\
    R060 & 2.4e-3 & 4.1e-5 & $\approx 58.5$ \\
    R100 & 3.2e-3 & 3.2e-5 & 100 \\
    R160 & 4e-3   & 2.5e-5 & 160 \\
    \rowcolor{tblhead}
    R250 & 5e-3   & 2e-5   & 250 \\
    R400 & 6.3e-3 & 1.6e-5 & $\approx 394$ \\
    R600 & 7.7e-3 & 1.3e-5 & $\approx 592$ \\
    R800 & 8.9e-3 & 1.1e-5 & $\approx 809$ \\
    \bottomrule
  \end{tabular}
\end{table}

\subsection{Motivation Experiments}
\begin{figure}[H]
  \centering
  \includegraphics[width=.4\textwidth]{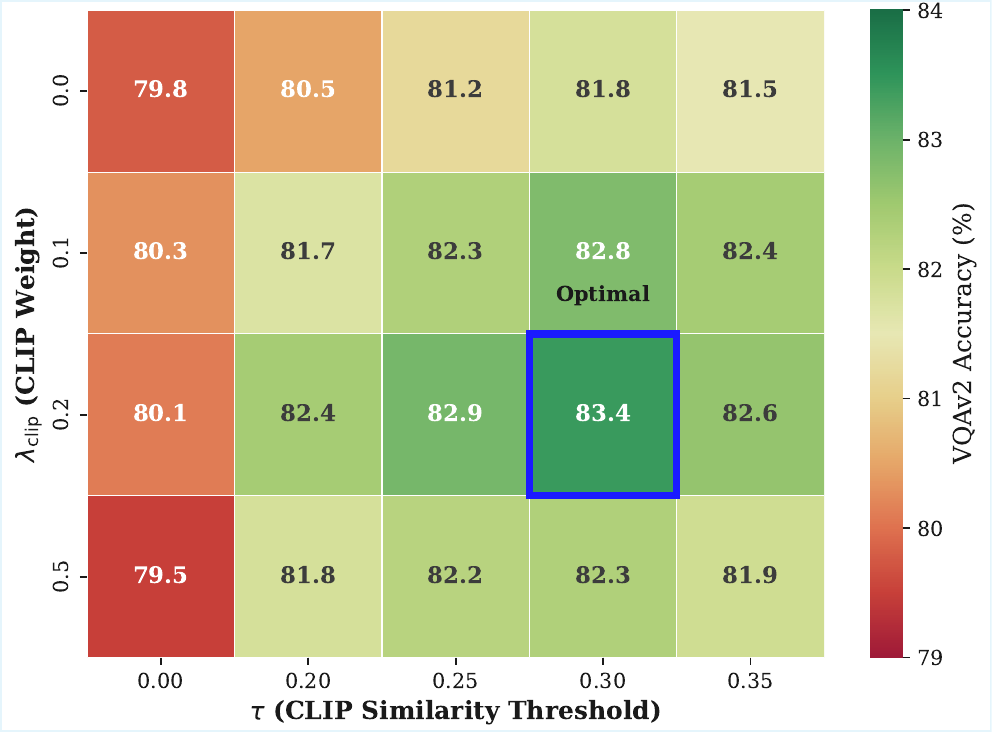}
  \caption{heatmap Ablation study on CLIP constraint configurations. We report VQAv2
        accuracy for different combinations of CLIP
        weight and CLIP similarity threshold}
  \label{fig:heatmap}
\end{figure}
\begin{figure}[t]
  \centering
  \includegraphics[width=.35\textwidth]{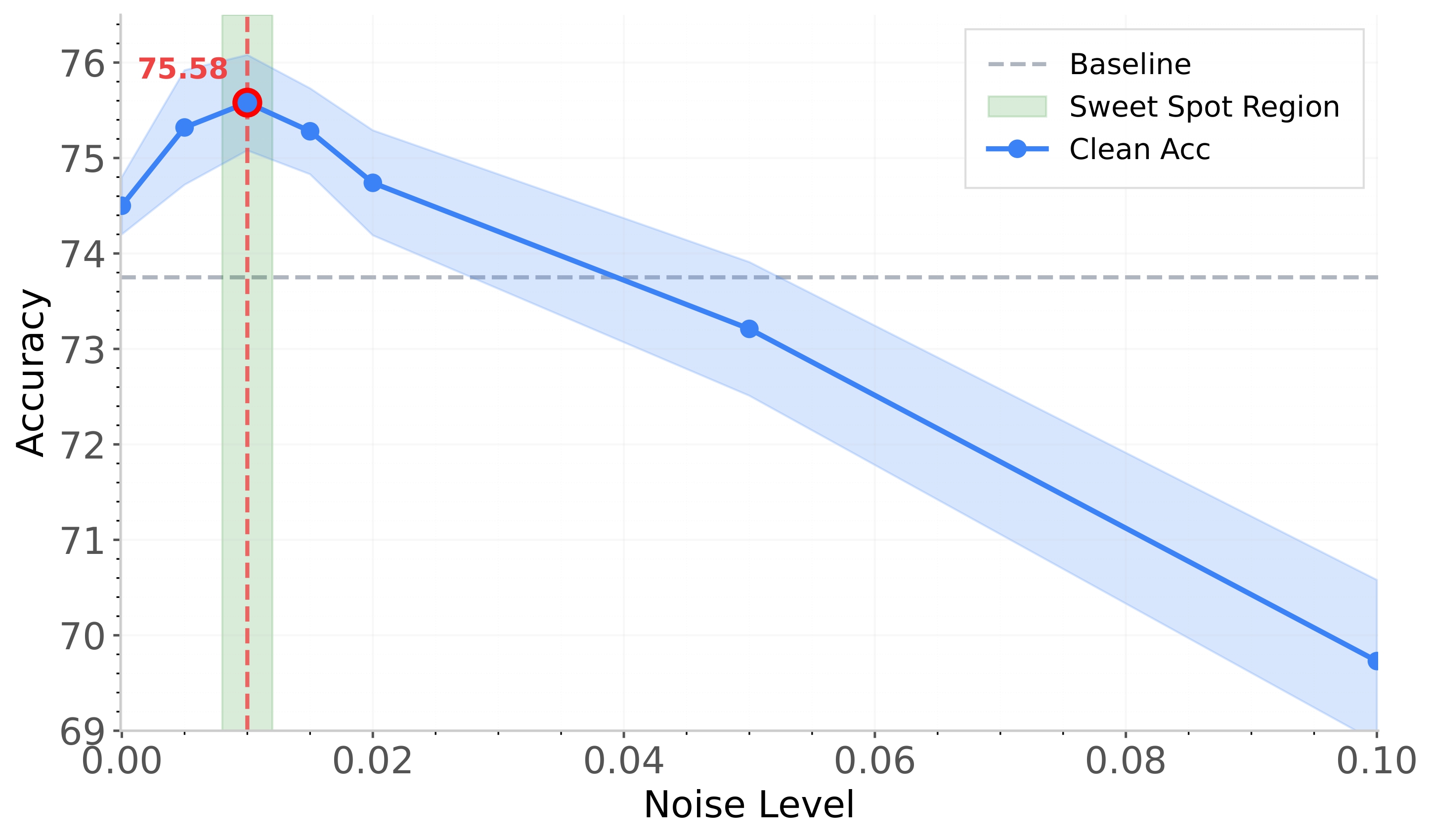}
  \caption{Perturbation Sweetspot}
  \label{fig:noisesweetspot}
\end{figure}
To find an Optimal noise level, we inject i.i.d.\ Gaussian noise into the projected visual tokens with $\sigma\in\{0, 0.005, 0.01, 0.015, 0.02, 0.05, 0.1\}$. 
We observe a \emph{sweet spot} near $\sigma\!\approx\!0.01$ where VQAv2 soft accuracy slightly increases (74.50\,$\to$\,75.58) before degrading at larger noise, see in \Cref{fig:noisesweetspot}. 
This indicates that small, structured embedding perturbations can beneficially modulate the shared representation.

\section{Robustness Results}
\label{sec-app-results-robustness}

The details results on OOD and adversarial robustness are shown in \Cref{tab:robustness_bench}, indicating that \method significantly improves the robustness of the models.

\begin{table}[htbp]
  \centering
  \caption{Results for OOD and adversarial robustness.}
  \label{tab:robustness_bench}
  \setlength{\tabcolsep}{6pt}
  \renewcommand{\arraystretch}{1.00}
  \footnotesize
  \begin{tabular}{
    l
    S[table-format=2.1] 
    S[table-format=2.1] 
  }
    \toprule
    \textbf{Model}
      & {\textbf{NaturalBench}}
      & {\textbf{AdVQA}} \\
    \midrule
    
    % InternLM-XC-V1           & 68.7 & \NA \\
    Janus-Pro  & 73.8 & 34.2 \\
    +SFT       & 73.9 & 36.4 \\
    \rowcolor{tblhilite}
    +Ours   & 78.6 & 40.4 \\
    \bottomrule
  \end{tabular}
\end{table}

\section{Details on Case Study}
\label{sec-append-case}

\subsection{Case Study on Understanding Tasks}
\label{sec-append-case-under}

We offer more interpretations to \Cref{fig:case_study_all}.
\begin{itemize}
    \item \textbf{Object counting} (C1): The baseline model fails to accurately count objects in cluttered scenes, often confusing similar-looking items or missing partially visible objects. 
After \method training, the model correctly identifies the precise count, demonstrating improved fine-grained visual attention.
\item \textbf{Object interaction} (C2): Understanding relational semantics between objects (e.g., "person holding umbrella" vs. "umbrella next to person") requires compositional reasoning. 
The baseline misinterprets spatial relationships, while \method correctly recognizes the interaction pattern.
\item \textbf{Spatial relation and location} (C3): Queries about relative positions (e.g., "left of", "behind") expose fragile spatial understanding in the baseline. 
\method's adversarial training—which systematically perturbs spatial layouts during decoding—hardens the model against such failures.
\item \textbf{Crowd object detection} (C4): dense and overlapping objects in crowded scenes challenge both localization and recognition. 
The baseline produces vague or incorrect answers, whereas \method maintains accuracy by learning from decoded adversarial samples that emphasize occlusion and clutter.
\end{itemize}

These qualitative improvements align with our quantitative gains, confirming that \method systematically addresses decision-critical reasoning failures rather than merely fitting to benchmark statistics.

In addition, we also present detailed analysis to the open-ended understanding tasks:
\begin{itemize}
  \item \textbf{Open-ended understanding.}
  As illustrated in \Cref{fig:casestudy}, \method produces more fine-grained and visually grounded captions than the baseline. The model not only recognizes the overall scene (e.g., pizza, street sign, animals) but also reliably captures details such as a missing pizza slice, vegetables on display at a farmers market, a cat sitting on a windowsill looking out the window, or two sheep wearing coats standing in a field. These examples show that adversarial self-play improves open-ended descriptions by encouraging the model to focus on decision-critical visual evidence rather than hallucinated or overly generic content.
\end{itemize}

\subsection{Case Study on Generation Tasks}
\label{sec-append-case-gen}

We offer more detailed explanation of the text-to-image generations in \Cref{fig:case_study_gen}.
\begin{itemize}
    \item On the synthetic shapes example, the baseline model already produces plausible objects but often violates fine-grained layout constraints (e.g., incorrect left/right ordering or cube–sphere counts), whereas \method yields images that respect the specified $2\times 2$ red cube stack, the correct number of blue spheres, and the spatial relations such as “on the left / on the right” and “between”.
    \item In the “broccoli in a glass bowl” example, \method more faithfully binds multiple attributes—three pieces of broccoli, two carrots on the side, and a clearly visible red sticker with the number “5” attached to the bowl—demonstrating stronger compositional control.
    \item For the Grand Canyon scene, the baseline sometimes collapses the layered rock formations into a flatter composition, while \method better preserves depth and lighting that match the prompt description.
    \item Finally, for the “blue-eyed Siamese cat sitting on a green velvet armchair”, \method produces a sharper Siamese appearance and a more coherent green velvet texture, indicating that self-play training can improve both semantic alignment and visual fidelity.
\end{itemize}

\section{Convergence and Hyperparameter Analysis}
\label{sec-append-conv}
\begin{figure}[t]
  \centering
  \includegraphics[width=.9\linewidth]{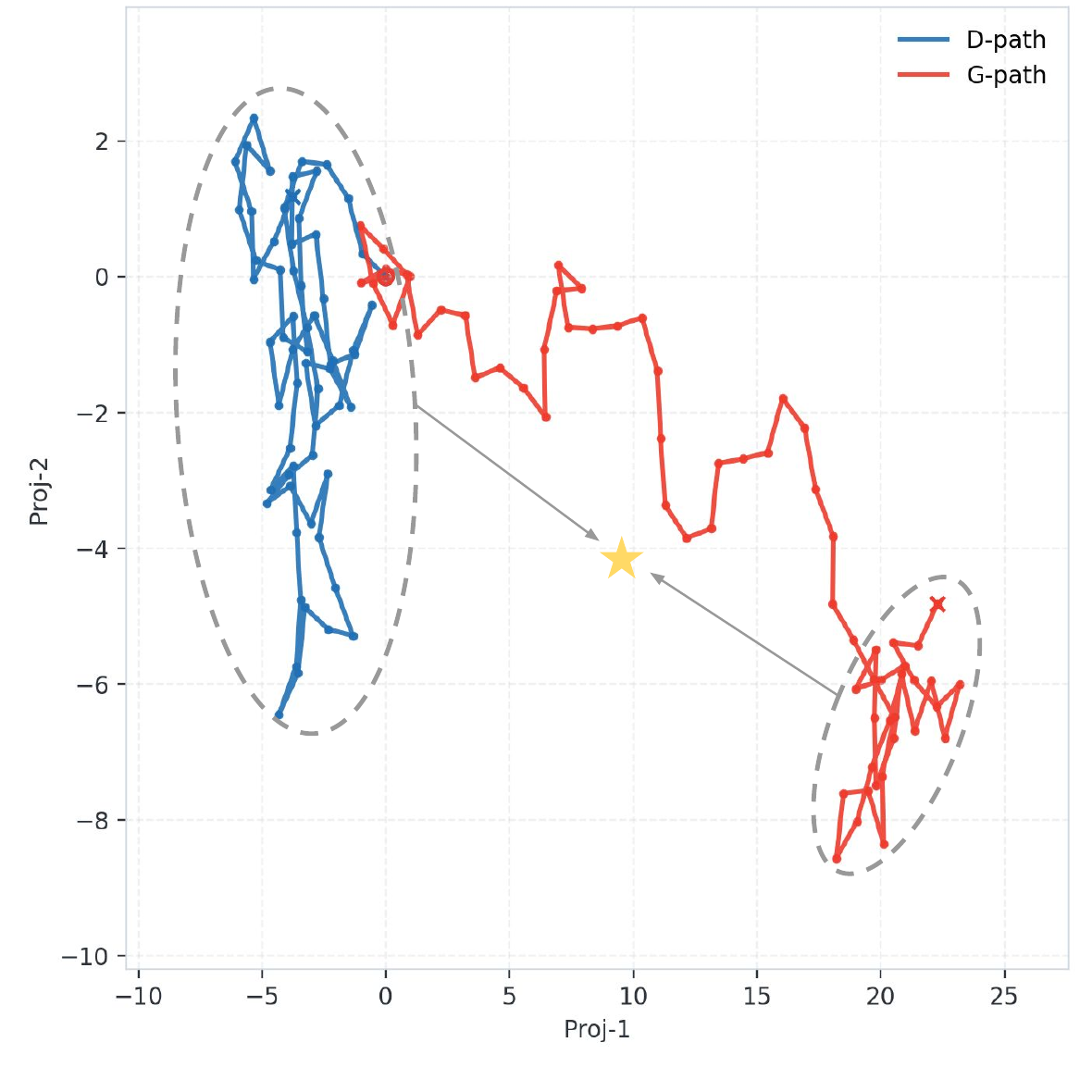}
  \caption{The best result of all of our runs, optimization path are projected to a two dimension axis.}
  \label{fig:opt-path}
\end{figure}

\subsection{Convergence}
\label{sec-append-conv-conv}

\begin{figure}[t]
    \includegraphics[width=.9\linewidth]{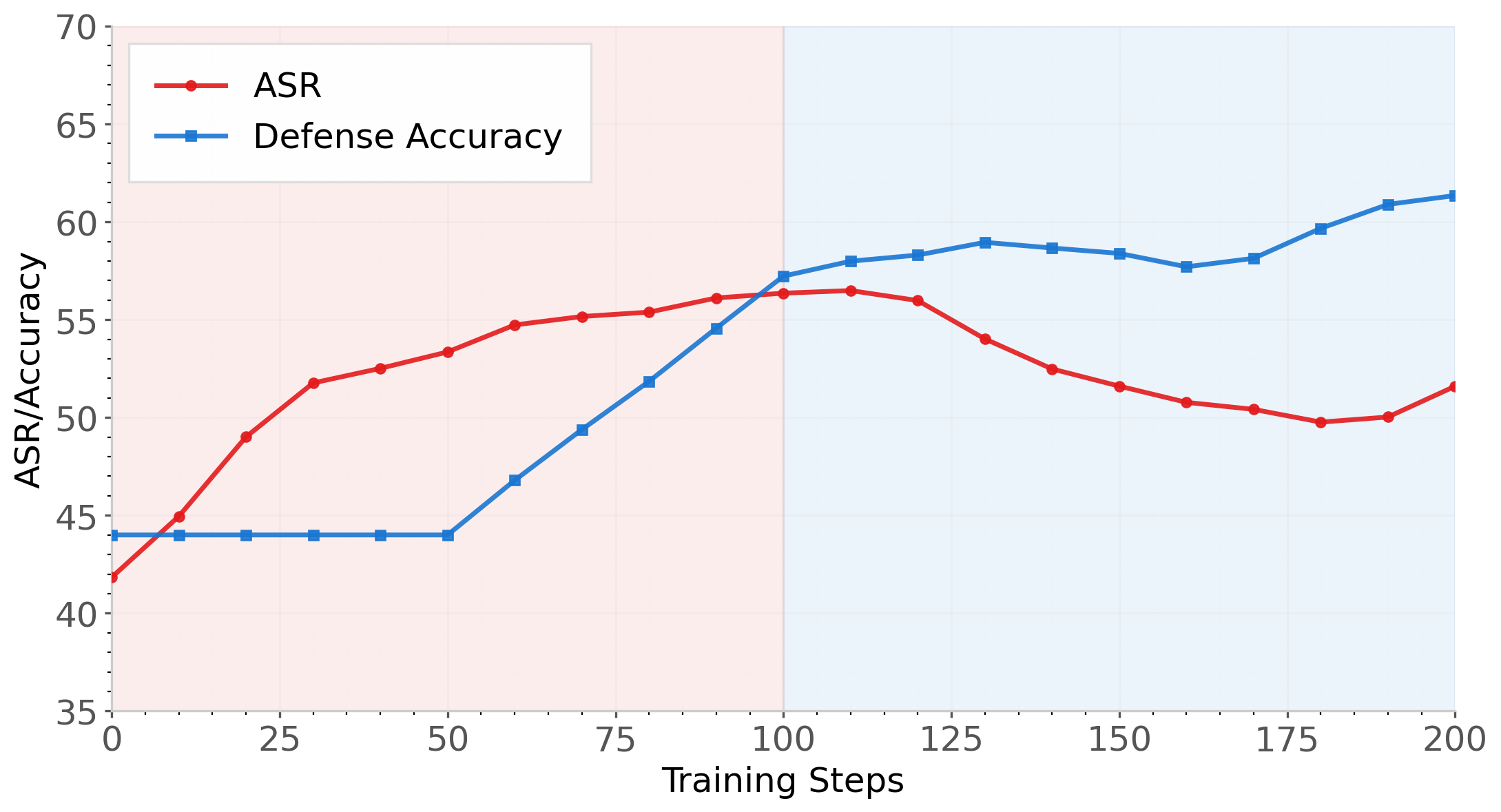}
    \caption{Self-play dynamics between the generation and the understanding . The two branches alternately dominate the training objective, exhibiting a stable tug-of-war behavior.}
    \label{fig:dynamic}
\end{figure}

\begin{figure}[t]
  \centering
  \includegraphics[width=\linewidth]{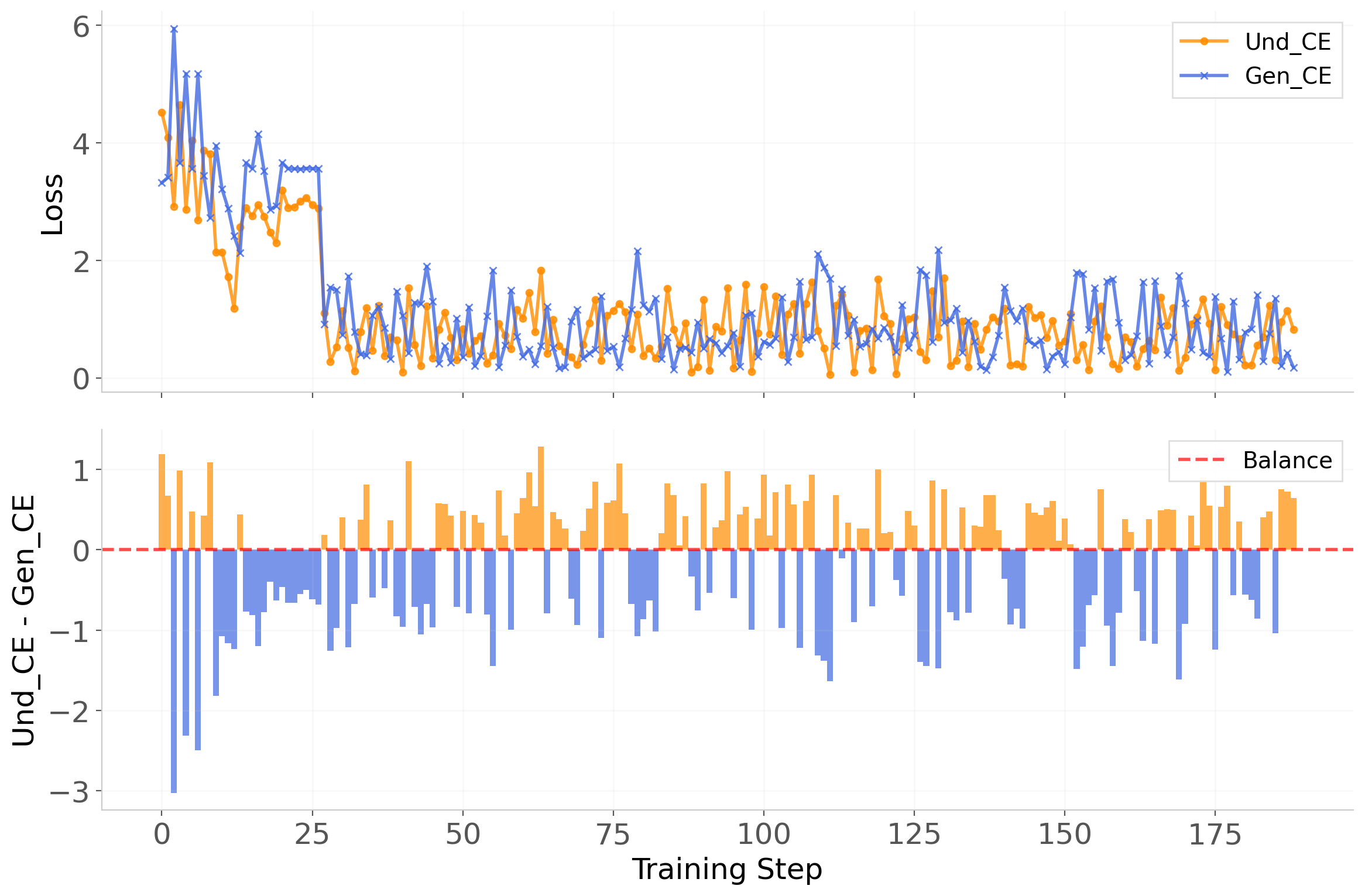}
  \caption{Dominance timeline. The trajectory alternates between understanding and generation phases, illustrating a stable tug-of-war rather than collapse to either side during training.}
  \label{fig:dominate-timeline}
\end{figure}

\noindent \textbf{Convergence of the minimax training.}
The minmax setup raises the practical question: when does the game converge and what schedules keep it stable?
In our setup, only the Perturber $C$ and LoRA adapters on the understanding branch $U$ are trainable; due to $U$'s larger capacity, it can dominate and degrade the generation module.
We restore stability by giving $C$ a higher learning rate and using short, interleaved updates.
We conducted an extensive sweep of the Generation/Understanding update ratio in \Cref{gd-ratio}, shows \(\texttt{gen\_lr}=5\times10^{-3}\), \(\texttt{und\_lr}=2\times10^{-5}\), provides the best clean–robust trade-off; prolonged generation phases saturate the attack success rate (ASR) before $U$ adapts and induce catch-up oscillations (see \Cref{fig:dominate-timeline} \Cref{fig:dynamic}).
Conversely, when the generation overpowers $U$, decoded candidates drift off-manifold and hurt clean accuracy.
Thus, \emph{balance progression speeds}: (i) use a slightly larger learning rate for $C$ than for $U$'s adapters, and (ii) prefer short alternations over long unilateral bursts.
Full grids, curves, and ablations are shown in \Cref{sec_app-experiment_details}.
\label{sec-append-convergence}
\noindent \textbf{Perturbation budget.}
The budget constraint $\varepsilon_{\mathrm{max}}$ controls the perturbation magnitude in the token space.
The results in \append{sec_app-experiment_details} show a sweetspot that inverted U-shaped performance curve \Cref{fig:noisesweetspot}, setting $\varepsilon_{\mathrm{max}}$ too small (e.g., $0.005$) produces weak perturbations that fail to expose critical reasoning failures, yielding limited robustness gains ($+1.7\%$ on NaturalBench). 

\subsection{Hyperparameter Sensitivity Analysis}
\label{sensitivity}

\begin{figure}[t]
    \includegraphics[width=\linewidth]{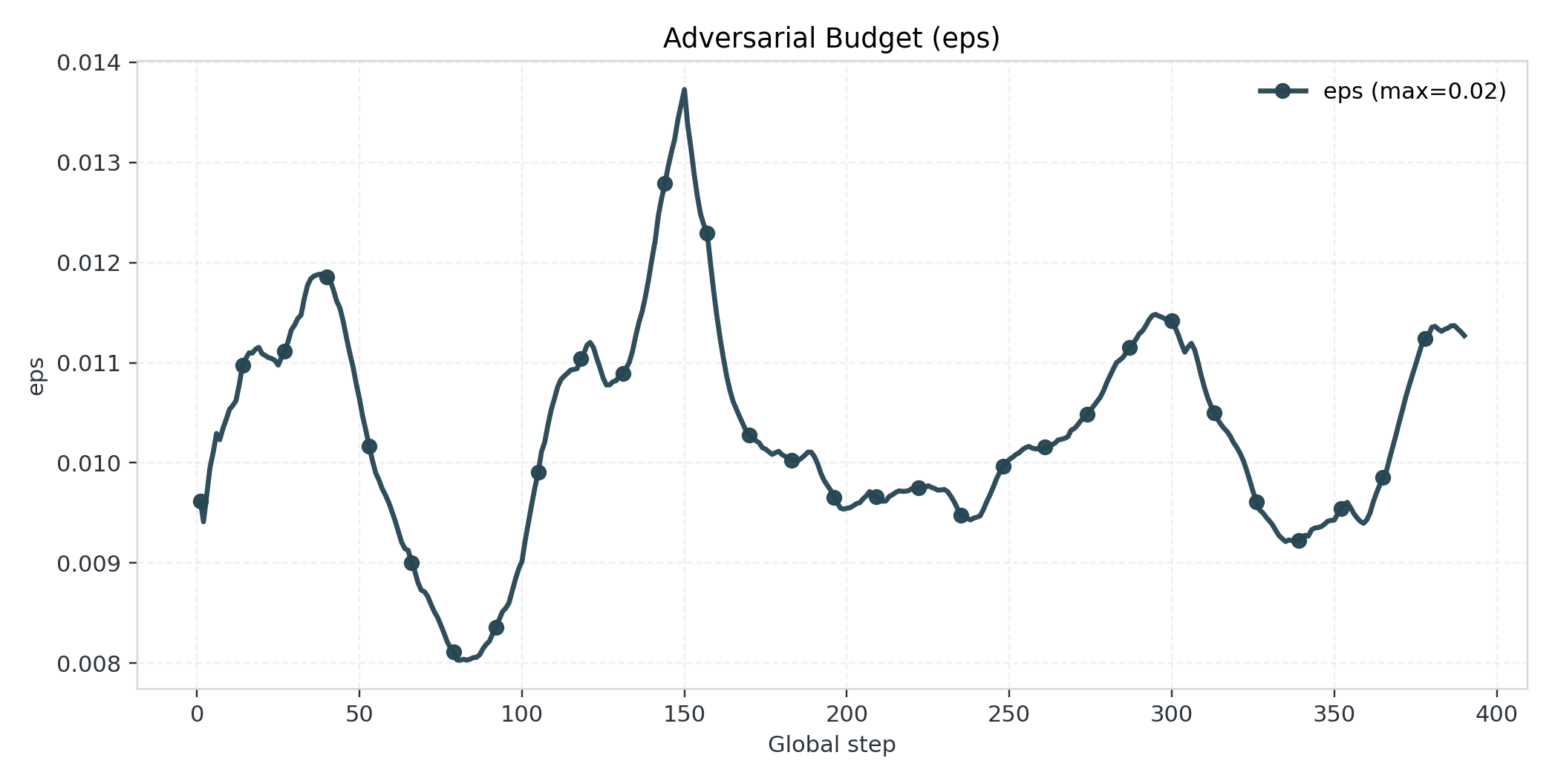}
    \caption{Perturebation Budget}
    \label{fig:eps_curve}
\end{figure}

Unless otherwise noted, we fix the perturbation budget to $\delta = \varepsilon_{\mathrm{max}} = 0.02$ in all main experiments, which we found to provide a good clean--robust trade-off after sweeping $\delta \in \{0.005, 0.01, 0.015, 0.02, 0.05, 0.10\}$ \Cref{fig:eps_curve}. 
For hard-example mining, we define the hardness score $H$ as the cross-entropy loss of the understanding branch on decoded candidates plus a CLIP-based hinge term, and select hard samples using a quantile-based threshold: the buffer threshold $\tau$ is set to the $60$-th percentile of $H$ within each mining batch, while additionally enforcing a minimum text--image CLIP similarity of $0.6$ to filter out semantically off-manifold generations. 
The trade-off coefficient $\beta$ in Eq.~\eqref{eq:lossU}, which weights the contribution of buffer samples relative to clean examples, is set to $\beta = 0.5$ by default so that roughly half of the understanding gradient comes from adversarial or hard instances; we observed that \method is numerically stable for a broad range of $\beta \in [0.3, 1.0]$. 
The hard-sample replay buffer stores up to $50$ decoded images ranked by $H$. We deliberately keep the capacity moderate, as substantially larger buffers (e.g., $\gg 10^4$ entries) would store many full-resolution decoded images and quickly lead to a steep increase in GPU and host memory usage, without providing noticeable additional benefits in practice.

\section{Theoretical Insights}
\label{sec:theory}
In this section, we provide preliminary theoretical justification for why the proposed minimax self-play procedure improves (i) the stability of the understanding branch, (ii) convergence of the alternating optimization, and (iii) coverage of the shared generative manifold. The analysis is intentionally model-agnostic and applies to a broad class of unified multimodal architectures.

\subsection{Convergence of the Minimax Self-Play Dynamics}

Recall the UniGame objective
\begin{equation}
\min_{\theta_U}\;\max_{\theta_C}\;
\mathcal{L}(\theta_U,\theta_C)
=
\mathbb{E}\big[ \ell_U(\theta_U) \big]
\;+\; 
\lambda\,\mathbb{E}\big[ \ell_C(\theta_C;\theta_U) \big],
\label{eq:minimax-objective}
\end{equation}
where the perturber maximizes the understanding loss while the understanding head minimizes both clean and adversarial losses, subject to a bounded perturbation $\|\delta\|\le\varepsilon_{\max}$ in the shared token space. In this subsection, we analyze an \emph{idealized} version of this minimax problem to provide theoretical intuition, rather than a full convergence proof for the actual deep network implementation.

\paragraph{Assumption 1 (Lipschitz continuity).}
The understanding loss $\ell_U(a \mid z, q)$ is $L$-Lipschitz continuous in the token embedding $z$ and continuously differentiable in $\theta_U$.

\paragraph{Assumption 2 (Bounded perturbation set and parameter domain).}
The perturber operates within a compact, convex set
\begin{equation}
\mathcal{D}=\{\delta:\|\delta\|\le\varepsilon_{\max}\}.
\end{equation}
Moreover, the parameter sets $\Theta_U$ and $\Theta_C$ for $\theta_U$ and $\theta_C$ are assumed to be compact and convex.

\paragraph{Assumption 3 (Local nonconvex--concave structure).}
For any fixed $\theta_U \in \Theta_U$, the function
$\theta_C \mapsto \mathcal{L}(\theta_U,\theta_C)$
is (locally) concave on $\Theta_C$ in a neighborhood of the stationary points of interest.
Equivalently, the game is nonconvex in $\theta_U$ and (locally) concave in $\theta_C$ around those points.

\paragraph{Proposition 1 (First-order stationary point and stability).}
Under Assumptions~1--3, the minimax problem in Eq.~\eqref{eq:minimax-objective}
admits at least one first-order stationary point $(\theta_U^\ast,\theta_C^\ast)$, i.e.,
\[
\nabla_{\theta_U}\mathcal{L}(\theta_U^\ast,\theta_C^\ast)=0,
\qquad
\nabla_{\theta_C}\mathcal{L}(\theta_U^\ast,\theta_C^\ast)=0.
\]
Moreover, for sufficiently small learning rates $(\eta_U,\eta_C)$, 
gradient descent--ascent generates a bounded sequence and converges to a 
neighborhood of a first-order stationary point of $\mathcal{L}$.

\paragraph{Sketch of proof.}
By Assumption~2, the feasible set in $(\theta_U,\theta_C,\delta)$ is compact and convex, so a minimax solution and hence a first-order stationary point exist.
Assumption~1 guarantees that the loss is smooth in $\theta_U$, and Assumption~3 provides a local nonconvex--concave structure: for each fixed $\theta_U$, the objective is (locally) concave in $\theta_C$.
Under such smooth nonconvex--concave conditions, standard results for two-player minimax optimization show that gradient descent--ascent with sufficiently small step sizes $(\eta_U,\eta_C)$ generates a bounded sequence and converges to an $\mathcal{O}(\eta_U + \eta_C)$ neighborhood of a first-order stationary point of $\mathcal{L}$.

\paragraph{Implication.}
These assumptions suggest that the adversarial self-play dynamics are stable and tend not to diverge, even though the perturber and understanding branches pursue opposing objectives.

\subsection{Robustness Improvement via Worst-Case Regularization}

For a fixed sample $z$ from the shared representation space, the creator seeks a worst-case perturbation
\begin{equation}
\max_{\|\delta\|\le\varepsilon_{\max}}
\ell_U(z+\delta).
\end{equation}
Using a first-order Taylor expansion around $z$, we obtain
\begin{equation}
\ell_U(z+\delta) 
\approx 
\ell_U(z) + \delta^\top \nabla_z\ell_U(z).
\label{eq:taylor}
\end{equation}
The optimal perturbation under the norm constraint is
\begin{equation}
\delta^\star 
= \varepsilon_{\max}
\,\frac{\nabla_z \ell_U(z)}{\|\nabla_z \ell_U(z)\|}.
\end{equation}
Substituting $\delta^\star$ into Eq.~\eqref{eq:taylor} and taking expectation over the data distribution yields the expected adversarial loss
\begin{equation}
\mathbb{E}\big[\ell_U(z) + 
\varepsilon_{\max}\|\nabla_z\ell_U(z)\|\big].
\end{equation}

\paragraph{Proposition 2 (Implicit gradient regularization).}
Adversarial self-play is equivalent, to first order, to adding a Jacobian-norm penalty:
\begin{equation}
\mathcal{L}_{U,\text{adv}} 
= \mathcal{L}_U
+\lambda\,\varepsilon_{\max} \,
\mathbb{E}\big[\|\nabla_z\ell_U(z)\|\big].
\label{eq:jacobian-penalty}
\end{equation}
Consequently, the understanding branch is encouraged to reduce its sensitivity to small perturbations in $z$, leading to locally flatter decision boundaries.

\paragraph{Implication.}
This explains the empirically observed improvements in robustness: the understanding head learns to be less sensitive to challenging input variations, improving both in-distribution and out-of-distribution performance as well as adversarial robustness.

\subsection{Manifold-Expanding Effect of Decoder-Constrained Perturbations}
\label{sec-append-theory-mani}

Unlike conventional pixel-space adversarial training, UniGame produces \emph{decoder-constrained} adversarial examples
\begin{equation}
\tilde{x}=G(z+\delta),\quad \tilde{x}\in \mathcal{M},
\end{equation}
where $G$ is the decoder and $\mathcal{M}$ is the decodable image manifold. This architecture ensures adversarial samples are:
\begin{enumerate}
    \item \textbf{On-manifold}: $\tilde{x}$ remains realistic and visually plausible;
    \item \textbf{Semantically valid}: filtered by CLIP-based or similar consistency criteria;
    \item \textbf{Near boundary regions}: targeted towards regions where the understanding model is fragile.
\end{enumerate}

\paragraph{Assumption 3 (Local bi-Lipschitz decoder).}
The decoder $G$ is locally bi-Lipschitz on the relevant region of the token space, i.e., there exist constants $0 < m \le M < \infty$ such that for all $z_1,z_2$ in a neighborhood $\mathcal{Z}$,
\begin{equation}
m \| z_1 - z_2 \|
\;\le\;
\|G(z_1) - G(z_2)\|
\;\le\;
M \| z_1 - z_2 \|.
\end{equation}

\paragraph{Lemma 1 (Adversarial manifold expansion).}
Under Assumption~3, for any $z \in \mathcal{Z}$ the support of the perturbed output distribution satisfies
\begin{equation}
\mathrm{supp}(G(z+\mathcal{D})) 
\supseteq \mathrm{supp}(G(z)),
\end{equation}
and expands the empirical training distribution toward regions where $\|\nabla_z \ell_U(z)\|$ is large.

\paragraph{Implication.}
The decoder-constrained perturbations induce a structured ``inflation'' of the data manifold towards decision boundary regions where the understanding head is uncertain. The hard-sample buffer $\mathcal{B}$ collects such samples, which are approximately located near the understanding decision boundary. Training on $\mathcal{B}$ reduces the empirical risk in these critical regions:
\begin{equation}
\hat{R}_{\text{adv}} = 
\frac{1}{|\mathcal{B}|}
\sum_{x\in\mathcal{B}} \ell_U(x)
\end{equation}
acts as a surrogate for minimizing the out-of-distribution risk $R_{\text{OOD}}$.

\subsection{Summary of Theoretical Insights}
The above analysis provides a theoretical lens on the benefits of the UniGame framework:
\begin{enumerate}
    \item \textbf{Convergence of self-play:} Alternating gradient descent--ascent admits a stationary saddle point under mild smoothness and compactness assumptions.
    \item \textbf{Robust optimization view:} The adversarial creator implicitly enforces a gradient-norm penalty (Eq.~\eqref{eq:jacobian-penalty}), flattening the understanding decision boundary.
    \item \textbf{Manifold expansion:} Decoder-constrained perturbations generate semantically valid hard samples that expand coverage of the decodable manifold towards challenging regions.
    \item \textbf{Alignment with empirical gains:} These properties theoretically support the empirical improvements in understanding, consistency, out-of-distribution robustness, and adversarial robustness observed in our experiments.
\end{enumerate}

\end{document}